\documentclass[letterpaper]{article} 
\usepackage{aaai21}  
\usepackage{times}  
\usepackage{helvet} 
\usepackage{courier}  
\usepackage[hyphens]{url}  
\usepackage{graphicx} 
\usepackage{natbib}  
\usepackage{caption} 
\usepackage{verbatim}
\usepackage{amssymb}
\usepackage{bbm}
\usepackage{bm}
\usepackage[tbtags]{amsmath}
\usepackage{amsthm}
\usepackage{enumitem}
\usepackage{subfigure}
\usepackage{booktabs}
\usepackage{nameref}
\usepackage{color}
\newtheorem{thm}{Theorem}
\newtheorem{lemma}{Lemma}
\newtheorem{defn}{Definition}[section]

\title{Margin-Based Transfer Bounds for Meta Learning with Deep Feature Embedding}
\author{
    Jiechao Guan\textsuperscript{\rm 1}\quad
    Zhiwu Lu\textsuperscript{\rm 1 \footnote{Corresponding author}}\quad
    Tao Xiang\textsuperscript{\rm 2}\quad
    Timothy Hospedales\textsuperscript{\rm 3}
    }
\affiliations{

    \textsuperscript{\rm 1}Beijing Key Laboratory
of Big Data Management and Analysis Methods, Gaoling School of
Artificial Intelligence, Renmin University of China\\
    \textsuperscript{\rm 2}Department of Electrical and Electronic Engineering,
University of Surrey\\
 \textsuperscript{\rm 3}School of Informatics,
The University of Edinburgh\\
abel\_guan@ruc.edu.cn\qquad luzhiwu@ruc.edu.cn
}
\begin{document}

\maketitle

\begin{abstract}
By transferring  knowledge learned from seen/previous tasks, meta learning aims to generalize well to unseen/future tasks. Existing meta-learning approaches have shown promising empirical performance on various multiclass classification problems, but few provide theoretical analysis on the classifiers' generalization ability on future tasks. In this paper, under the assumption that all classification tasks are sampled from the same meta-distribution, we leverage margin theory and statistical learning theory to establish three margin-based transfer bounds for meta-learning based multiclass classification (MLMC). These bounds reveal that the expected error of a given classification algorithm for a future task can be estimated with the average empirical error on a finite number of previous tasks, uniformly over a class of preprocessing feature maps/deep neural networks (i.e. deep feature embeddings). To validate these bounds, instead of the commonly-used cross-entropy loss, a multi-margin loss is employed to train a number of representative MLMC models. Experiments on three benchmarks show that these margin-based models still achieve competitive performance, validating the practical value of our margin-based theoretical analysis.
\end{abstract}

\section{Introduction}

Inspired by human's ability to recognizing an unseen/new object category, meta-learning based multiclass classification (MLMC), one instantiation of which is $k$-way $s$-shot classification (e.g., $k=5$ and $s=10$) \cite{oriol2016nips}, has been  studied intensively in the past few years. It is often cast into a meta-learning scenario \cite{thrun1998springer}, in which a \emph{meta-learner} learns prior knowledge from several training tasks and then facilitates a \emph{base-learner} to generalize well on unseen/future tasks. Recent meta-learning based classification models normally employ a deep convolutional neural network to learn each task. As a typical example, the meta-learner learns a prior which is often in the form of a feature extractor (i.e., the lower layers of the network, often called as deep feature embedding), and the base-learner relies on the prior (with the feature extractor frozen) to update the fully connected layer for classification given a new task. The goal of MLMC is thus to learn optimal deep feature embedding from observed tasks and train a new classifier which performs well on a future task.

Existing meta-learning based classification approaches \cite{oriol2016nips, jake2017nips, sung2018cvpr, finn2017icml} have shown promising empirical performance on several benchmarks, but few provide theoretical analysis on the expected performance of their learned classifiers on new/unseen tasks, which may have different data distributions from that of previous training tasks. To overcome this limitation, the focus of this paper is thus on providing theoretical guarantees on how well a meta-learning based classification model can generalize to new/unseen tasks.

The central assumption of our theoretical results is that the learner is embedded within a distribution of related learning tasks. Since all learning tasks, either from seen classes or unseen ones, share some similarities (e.g., all tasks on the CUB dataset \cite{WahCUB_200_2011} are about recognizing fine-grained bird species), it is reasonable to assume that these tasks are sampled from a common meta-distribution, which is referred to as \emph{environment} in \cite{baxter2000jai}. Given that each task represents a data distribution, an environment can be considered as a distribution of distributions. Starting with a steam of datasets drawn from different training tasks in this environment, we aim to learn a multiclass classifier which minimizes the \emph{transfer risk} \cite{baxter2000jai, maurer2009ml, maurer2016jmlr} for a new task randomly sampled from the same environment.

Under this assumption, we leverage the margin theory \cite{vapnik1982springer} and the statistical learning theory \cite{valiant1984acm} to derive a high probability bound of transfer risk. The margin theory has been utilized to study  behaviours of many machine learning models \cite{schapire1997icml, koltchinskii2002annual_stat, bartlett2017nips}, and is a standard tool to analyze multiclass classification problems (see Chapter~9 in \cite{mohri2012book}). Under the \emph{probably approximately correct} (PAC) learning framework, we establish a margin-based transfer bound with Gaussian complexity \cite{bartlett2002jmlr} for MLMC (see Theorem~\ref{thm:main_gaussian_complexity}). Bounding the Gaussian complexity with chaining \cite{dudley1967jfa} and other statistical learning techniques, we further provide a margin-based transfer bound with covering number \cite{anthony2002theory} (in Theorem~\ref{thm:main_covering_number}) and a transfer bound based on the VC-dimension \cite{vaart1996springer} of the given scoring function class (in Theorem~\ref{thm:transfer_bound_fsl}), respectively. These results demonstrate that, for any fixed preprocessing deep neural networks (meta-learner) and any given classification algorithm (base-learner), the expected error for a future MLMC task can be controlled by the average empirical margin loss on the training tasks.
This theoretical analysis is applicable to any MLMC methods which do not update the parameters of the meta learner/feature extractor when performing classification in a new/unseen task. It thus covers a quite wide range of methods (see examples of such MLMC variants in Sect.~\nameref{sect:experiments}).

Our main contributions are three-fold: \vspace{0.05cm}\\
\textbf{(1)} Our main theorem (i.e. Theorem~\ref{thm:transfer_bound_fsl}) gives a rigorous theoretical statement that an MLMC model's transfer risk on an unseen task can be bounded by the empirical error on previous tasks plus a complexity part. This transfer bound guarantees that under certain constraints (e.g., the scoring function class's VC-dimension $v$ is finite and the task number $n$ becomes large), the average empirical margin loss is a proper estimation of the expected loss on a new task. Further, our main theorem also reveals that the obtained transfer bound admits only a linear dependency on the number $k$ of classification categories. To obtain this transfer bound, we first provide the margin-based transfer bounds for MLMC with Gaussian complexity and covering number, in Theorem~\ref{thm:main_gaussian_complexity} and Theorem~\ref{thm:main_covering_number}, respectively. \vspace{0.05cm}\\
\textbf{(2)} Our transfer bounds (e.g. Theorem~\ref{thm:transfer_bound_fsl}) are all dimension free when deep feature embedding is used for MLMC. Importantly, for a meta-learning problem in which a meta-learner learns a neural network from previous tasks and a base-learner learns a new classifier to generalize on a future task, our theoretical results reveal that the significant step is to learn a proper deep feature embedding function (combined with any multiclass classification algorithm) which can induce a large family of classifiers containing a good solution for any task sampled from the same environment. In addition, for meta learning with deep feature embedding, the sample efficiency per task can actually be guaranteed by our theoretical results (see more detailed discussion at the end of the related work section). \vspace{0.05cm}\\
\textbf{(3)} We adopt the multi-margin loss (the surrogate of the margin loss), rather than the cross-entropy loss commonly used in multiclass classification problems, to train existing typical MLMC models. The experimental results (see Tables~\ref{tab:main_mini}--\ref{tab:cross_domain_fsl} in Sect.~\nameref{sect:experiments}) demonstrate that the trained models with margin loss still achieve competitive performance on three benchmark datasets (miniImageNet \cite{oriol2016nips}, CUB \cite{WahCUB_200_2011} and miniImageNet$\rightarrow$CUB). This clearly validates the practical value of our margin-based theoretical analysis for MLMC. \vspace{0.05cm}

The remainder of this paper is organized as follows. Sect.~\nameref{sect:preliminary} provides the background information and notations. Sect.~\nameref{sect:our_results} presents our main theoretical results on the transfer bound of MLMC, followed by the empirical experiments in Sect.~\nameref{sect:experiments}. The differences between our theoretical results and closely-related works are discussed in Sect.~\nameref{sect:related_work}. Sect.~\nameref{sect:conclusions} draws conclusions and points out future research directions, respectively.

\section{Preliminary}
\label{sect:preliminary}

\subsection{Learning Setup}
\label{subsect:learning_setup}

\textbf{Tasks and Samples}. In meta-learning multiclass classification (MLMC) problem, a task is a probability measure $\mu \in M_1(\mathcal{X} \times \mathcal{Y})$, where $\mathcal{X}$ is an input space, $\mathcal{Y}$ is an output space (which is $\{1,...,k\}$ in multiclass classification), and $M_1(\mathcal{H})$ generally denotes the set of probability measures on a space $\mathcal{H}$. Information about the task $\mu$ is obtained by independently sampling a finite number $m$ of training examples $(x_i, y_i) \sim \mu$. Such an $m$-tuple $((x_1, y_1),...,(x_m, y_m)) \sim \mu^{m}$ is called a \emph{sample}, which is also denoted as $(\mathbf{x},\mathbf{y})=((x_1, y_1),...,(x_m, y_m))$ with $\mathbf{x}=(x_1,...,x_m)$ and $\mathbf{y}=(y_1,...,y_m)$.\\

\noindent\textbf{Algorithms}. We consider the classification problem with the hypothesis space $\mathcal{F}$ of scoring functions. A meta-learning classification algorithm (e.g., based on SVM) is a function $f$: $(\mathcal{X} \times \mathcal{Y})^{m} \rightarrow \mathcal{F}$. Let $\mathcal{D}$ be a space of alternative feature maps/neural networks, and for deep learning based methods, we need to choose a feature map $\varphi~(\in \mathcal{D})$ to preprocess the input data. This induces a new MLMC classification algorithm $f_{\varphi}: (\mathcal{X} \times \mathcal{Y})^{m} \rightarrow \mathcal{F}$. We can regard $\varphi$ and $f$ as the meta-learner and the base-learner, respectively. From the training sample $(\mathbf{x},\mathbf{y})$, this algorithm learns a scoring function $f_{\varphi(\mathbf{x}, \mathbf{y})} \in \mathcal{F}$: for each input-output pair $(x_{i}, y_{i})$, the scoring function outputs the prediction score $f_{\varphi(\mathbf{x}, \mathbf{y})}(x_{i}, y)$, i.e., the probability of $x_i$ belonging to the class label $y$ ($y \in \mathcal{Y}$). Without loss of generality, we assume that there exits a positive number $b>0$, such that $|f_{\varphi(\mathbf{x}, \mathbf{y})}(x,y)| \leq b,$ $\forall f_{\varphi(\mathbf{x}, \mathbf{y})} \in \mathcal{F}, (x,y) \sim \mu$.\\

\noindent\textbf{Environments}. The encounter with a task $\mu$ is itself a random event, corresponding to a draw $\mu \sim \varepsilon$ where $\varepsilon$ is a probability measure on the set of tasks, i.e., $\varepsilon \in M_1(M_1(\mathcal{X} \times \mathcal{Y}))$. In this work, such probability measures are called $\emph{environments}$ as in \cite{baxter2000jai, maurer2009ml}. Information about the environment $\varepsilon$ is obtained by independently drawing a finite number $n$ of tasks $\{\mu_{l}:\mu_{l}\sim \varepsilon,l=1,...,n\}$: each task $\mu_{l}$ is represented by a sample $(\mathbf{x}^{l}, \mathbf{y}^{l})\sim (\mu_{l})^{m}$, $(\mathbf{x}^{l}, \mathbf{y}^{l})=((x_1^l,y_1^l),...,(x_m^l, y_m^l))$, with the understanding that $\mathbf{x}^{l}=(x_1^l, ..., x_m^l)$ and $\mathbf{y}^{l}=(y_1^l, ..., y_m^l)$. Under the MLMC setting, the size $m$ of each task is kept the same to facilitate the analysis. Let $(\mathbf{X}, \mathbf{Y})=((\mathbf{x}^{1},\mathbf{y}^{1}), ..., (\mathbf{x}^{n}, \mathbf{y}^{n}))$ be the training data generated in this manner. Further, we define a probability measure $\hat{\varepsilon}$ on the set of samples $(\mathcal{X} \times \mathcal{Y})^{m}$ by letting expectation $\mathbb{E}_{\hat{\varepsilon}}(g)= \mathbb{E}_{\mu \sim \varepsilon}\mathbb{E}_{(\mathbf{x},\mathbf{y})\sim \mu^{m}}g(\mathbf{x},\mathbf{y})$ for every Borel measurable function $g$ on $(\mathcal{X} \times \mathcal{Y})^{m}$. The entire training data $(\mathbf{X}, \mathbf{Y})$ can thus be considered to be generated in $n$ independent draws from $\hat{\varepsilon}$, that is, $(\mathbf{X}, \mathbf{Y}) =((\mathbf{x}^{1},\mathbf{y}^{1}), ..., (\mathbf{x}^{n}, \mathbf{y}^{n})) \sim {\hat{\varepsilon}}^{n}$.

\subsection{Margin Loss and Transfer Risk}
\label{subsect:margin}
The margin of a scoring function $f_{\varphi(\mathbf{x}, \mathbf{y})}$ (trained from the sample $(\mathbf{x}, \mathbf{y})$ ) at a labeled data $(x_i,y_i)$ is
\begin{equation*}
\rho_{f_{\varphi(\mathbf{x}, \mathbf{y})}}(x_i,y_i)\triangleq f_{\varphi(\mathbf{x}, \mathbf{y})}(x_i,y_i)-\max_{y\neq y_i}f_{\varphi(\mathbf{x}, \mathbf{y})}(x_i, y).
\end{equation*}
A real-valued function associated with any algorithm $f_{\varphi}$ on the training sample $(\mathbf{x}, \mathbf{y})$ is its empirical loss $\hat{\ell}_{f_{\varphi}}: (\mathcal{X} \times \mathcal{Y})^{m} \rightarrow \mathbb{R}^{+}$, defined by
\begin{equation*}
\hat{\ell}_{f_{\varphi}}(\mathbf{x}, \mathbf{y})= \frac{1}{m}\sum_{i=1}^{m}\ell_{f_{\varphi(\mathbf{x}, \mathbf{y})}}(x_i, y_i)
\label{eq:emp_loss}
\end{equation*}
where $\ell_{f_{\varphi(\mathbf{x}, \mathbf{y})}}(x_i, y_i) = \Phi_{\rho}{\scalebox{.8}{$\circ$}}\rho_{f_{\varphi(\mathbf{x}, \mathbf{y})}}(x_i, y_i)$ and ${\scalebox{.8}{$\circ$}}$ denotes the operator of function composition, and for any $\rho >0$, the margin loss $\Phi_{\rho}(x)=\min\big(1, \max(0, 1-\frac{x}{\rho})\big)$. To evaluate the performance of a MLMC algorithm $f_{\varphi}$ ($\varphi \in \mathcal{D}$) in an environment $\varepsilon$,  the following steps are taken: (\romannumeral 1) make a random choice of a task $\mu \sim \varepsilon$, (\romannumeral 2) draw a training sample $(\mathbf{x}, \mathbf{y}) \sim \mu^{m}$, (\romannumeral 3) select a test pair $(x,y) \sim \mu$, (\romannumeral 4) run the algorithm $f_{\varphi}$ to obtain scoring function $f_{\varphi(\mathbf{x},\mathbf{y})}$, (\romannumeral 5) return the loss $\ell_{f_{\varphi(\mathbf{x},\mathbf{y})}}(x,y)$. The expected output of this procedure can be used to measure the generalization ability of the MLMC algorithm in the given environment. This motivates the following definition of the expected \emph{transfer risk} associated with the learning algorithm $f_{\varphi}$:
\begin{equation}
R_{\varepsilon}(f_{\varphi})=\mathbb{E}_{\mu \sim \varepsilon}\mathbb{E}_{(\mathbf{x},\mathbf{y}) \sim \mu^{m}}\mathbb{E}_{(x,y) \sim \mu}\ell_{f_{\varphi(\mathbf{x},\mathbf{y})}}(x,y)
\label{eq:transfer_loss}.
\end{equation}
Given all training data $(\mathbf{X},\mathbf{Y}) \sim \hat{\varepsilon}^{n}$, we utilize $(\mathbf{X}, \mathbf{Y})$ to select a $\varphi(\mathbf{X}, \mathbf{Y}) \in \mathcal{D}$ and fix it on the future task, so that the expected transfer risk $R_{\varepsilon}(f_{\varphi(\mathbf{X}, \mathbf{Y})})$ of the modified algorithm $f_{\varphi(\mathbf{X}, \mathbf{Y})}$ is minimal or near minimal. The conceptually simplest way is to select $\varphi(\mathbf{X}, \mathbf{Y}) = \arg \min_{\varphi \in \mathcal{D}}\frac{1}{n}\sum_{l=1}^{n}\hat{\ell}_{f_{\varphi}}(\mathbf{x}^{l}, \mathbf{y}^{l})$, which minimizes the average empirical risk on the available training data. In this paper, we give a high probability bound of $R_{\varepsilon}(f_{\varphi})$ in terms of $\frac{1}{n}\sum_{l=1}^{n}\hat{\ell}_{f_{\varphi}}(\mathbf{x}^{l}, \mathbf{y}^{l})$, and such bound uniformly holds for all $ \varphi \in \mathcal{D}$, not just for $\varphi(\mathbf{X}, \mathbf{Y})$ (see Theorem~\ref{thm:transfer_bound_fsl}).

\subsection{VC-dimension and Gaussian Complexity}

\begin{defn}[VC-dimension, 2.6.1 in \cite{vaart1996springer}]
Let $\mathcal{C}$ be a collection of subsets of a set $\mathcal{X}$. $\mathcal{C}$ is said to \emph{shatter} $\{x_1,...,x_n\}$ if each of its $2^n$ subsets can be expressed as the form $C \bigcap \{x_1,...,x_n\}$ for a $C$ in $\mathcal{C}$. The VC-dimension of the class $\mathcal{C}$ is the largest $n$ for which a set of size $n$ is shattered by $\mathcal{C}$.
\end{defn}
\begin{defn}[VC-dimension of Real-Valued Function Class]
The \emph{subgraph} of a function $f (\in \mathcal{F}): \mathcal{X}\rightarrow \mathbb{R}$ is the subset of $\mathcal{X} \times \mathbb{R}$ given by $\{(x,t):t < f(x)\}$. Then the VC-dimension of the function class $\mathcal{F}$ is defined as the VC-dimension of the set of subgraphs of functions in $\mathcal{F}$.
\end{defn}
\begin{defn}[Gaussian Complexity \cite{bartlett2002jmlr}]
For a subset $A \subseteq \mathbb{R}^{m}$, the Gaussian complexity of $A$ is defined as $\Gamma(A)= \mathbb{E}_{\gamma}\sup_{\mathbf{x} \in A}2/m\sum_{i=1}^{m}\gamma_i x_i$, where $\{\gamma_i\}_{i\ge 1}$ is a sequence of independent standard Gaussian variables (i.e. $\gamma_i \sim N(0,1)$).
\end{defn}
If $\mathcal{F}$ is a class of real-valued functions on the space $\mathcal{X}$ and $\mathbf{x}=(x_1,...,x_m) \in \mathcal{X}^{m}$, we define
$\mathcal{F}(\mathbf{x})=\mathcal{F}(x_1,...,x_m)  = \{(f(x_1),...,f(x_m)):f \in \mathcal{F}\} \subseteq \mathbb{R}^{m}.$
The empirical Gaussian complexity of $\mathcal{F}$ on $\mathbf{x}$ is $\Gamma(\mathcal{F}(\mathbf{x}))$. Let $\mu \in M_1(\mathcal{X})$ be a probability measure on $\mathcal{X}$, and the corresponding expected complexity is $\mathbb{E}_{(\mathbf{x}, \mathbf{y}) \sim \mu^{m}}\Gamma(\mathcal{F}(\mathbf{x}))$. $\mathcal{F}(\mathbf{X})$ and its expected Gaussian complexity $\mathbb{E}_{(\mathbf{X}, \mathbf{Y}) \sim \hat{\varepsilon}^{n}} \Gamma(\mathcal{F}(\mathbf{X}))$ can be defined in a similar way.

\section{Theoretical Results}
\label{sect:our_results}

In this section, we present our theoretical results. The most important one is Theorem~\ref{thm:transfer_bound_fsl}, which reveals that the learning bound of an algorithm on the new MLMC tasks (which may have different data distributions from that of previous tasks) can be controlled by the empirical loss on previous tasks plus the complexity term. \textbf{All detailed proofs for our theoretical results can be found in the supplementary material}.

\begin{thm}[\textbf{Margin-Based Transfer Bound for MLMC with VC-dimension}]
Assume that the VC-dimension of the real-valued function class $\Pi_1\mathcal{F} = \{x \mapsto g(x,y) | g \in \mathcal{F}, y \in \mathcal{Y}, |\mathcal{Y}|=k\}$ is $v$, and $\Pi_1\mathcal{F}$ is uniformly bounded by $b>0$. Given a classification algorithm $f$ and a margin parameter $\rho >0$, for any environment $\varepsilon \in M_1(M_1(\mathcal{X} \times \mathcal{Y}))$ and for any $\delta > 0$, with probability at least $1-\delta$ on the data $(\mathbf{X}, \mathbf{Y}) \sim {\hat{\varepsilon}}^{n}$, we have for all feature maps $\varphi \in \mathcal{D}$ that
\begin{equation*}
\begin{split}
R_{\varepsilon}(f_{\varphi})\leq & \frac{1}{n}\sum_{l=1}^{n}\hat{\ell}_{f_{\varphi}}(\mathbf{x}^{l}, \mathbf{y}^{l}) + \sqrt{\frac{\ln(1/\delta)}{2n}}\\
+  & (\frac{k}{\rho\sqrt{m}}+\frac{k}{\rho\sqrt{n}})(C_1 \sqrt{v} + C_2),
\end{split}
\label{thm:transfer_bound_fsl}
\end{equation*}
where constants $C_1 = 24\sqrt{2\pi}b(1+\sqrt{\log (16e)}+2\sqrt{2})$ and $C_2 = 24\sqrt{2\pi}b(\sqrt{\log C_0} + \sqrt{\log (16e)})$. $C_0$ is the uniform constant defined in Theorem~\ref{thm:covering_number_vc_bound}.
\end{thm}
The main proof is based on Theorem~\ref{thm:main_covering_number}. To obtain Theorem~\ref{thm:main_covering_number}, we first give the Gaussian complexity transfer bound in Theorem~\ref{thm:main_gaussian_complexity}, which is accomplished by using Slepian's Lemma \cite{Ledoux1991springer} to bound the function class $\mathcal{G}_{\varphi}=\{(\mathbf{x}, \mathbf{y}) \mapsto \hat{\ell}_{f_{\varphi}}(\mathbf{x}, \mathbf{y})\}$. This is not straightforward and thus contributes our major technical novelty.

\subsection{Gaussian Complexity Transfer Bound for MLMC}
\label{subsect:gaussian_complexity_bound}

\begin{thm}[\textbf{Margin-Based Transfer Bound for MLMC with Gaussian Complexity}]
Let $\mathcal{F}$ be a hypothesis of scoring functions. Given a classification algorithm $f$ and a margin parameter $\rho >0$, for any environment $\varepsilon \in M_1(M_1(\mathcal{X} \times \mathcal{Y}))$ and for any $\delta > 0$, with probability at least $1-\delta$ on the data $(\mathbf{X}, \mathbf{Y}) \sim {\hat{\varepsilon}}^{n}$, we have for all feature maps $\varphi \in \mathcal{D}$ that
\begin{equation*}
\begin{split}
R_{\varepsilon}(f_{\varphi})\leq & \hspace{-0.01in}\frac{1}{n}\sum_{l=1}^{n}\hspace{-0.01in}\hat{\ell}_{f_{\varphi}}(\mathbf{x}^{l}, \mathbf{y}^{l}) +  \frac{k\sqrt{2m\pi}}{\rho}\hspace{-0.1in}\mathop{\mathbb{E}}\limits_{(\mathbf{X},\mathbf{Y}) \sim \hat{\varepsilon}^{n}}\hspace{-0.1in}\Gamma(\Pi_{1}\mathcal{F}(\mathbf{X}))  \\
 + & \sqrt{\frac{\ln(1/\delta)}{2n}} + \frac{k\sqrt{2\pi}}{\rho}\hspace{-0.02in}\mathop{\mathbb{E}}\limits_{\mu \sim \varepsilon}\mathop{\mathbb{E}}\limits_{(\mathbf{x}, \mathbf{y}) \sim \mu^{m}}\hspace{-0.1in}\Gamma(\Pi_1\mathcal{F}(\mathbf{x})),
\end{split}
\end{equation*}
where $\Pi_1\mathcal{F}(\mathbf{X})=\{\big(f_{\varphi_{(\mathbf{X},\mathbf{Y})}}(x_{1}^{1},y),...,f_{\varphi_{(\mathbf{X},\mathbf{Y})}}(x_{m}^{1},y),\\
...,f_{\varphi_{(\mathbf{X},\mathbf{Y})}}(x_{1}^{n},y),...,f_{\varphi_{(\mathbf{X},\mathbf{Y})}}(x_{m}^{n},y)\big): y \in \mathcal{Y}, \varphi \in \mathcal{D}\}$, $\Pi_1\mathcal{F}(\mathbf{x})=\{\big(f_{\varphi(\mathbf{x},\mathbf{y})}(x_{1},y),...,f_{\varphi(\mathbf{x},\mathbf{y})}(x_{m},y)\big): y \in \mathcal{Y}, \varphi \in \mathcal{D}\}$, and the scoring function $f_{\varphi_{(\mathbf{X},\mathbf{Y})}}$ is defined as: $f_{\varphi_{(\mathbf{X},\mathbf{Y})}}(x_{i}^{l}, y) = f_{\varphi(\mathbf{x}^{l},\mathbf{y}^{l})}(x_{i}^{l},y), \forall i \in [m], l \in [n]$.
\label{thm:main_gaussian_complexity}
\end{thm}

The main proof strategy is to rewrite $R_{\varepsilon}(f_{\varphi}) - \frac{1}{n}\sum_{l=1}^{n}\hat{\ell}_{f_{\varphi}}(\mathbf{x}^{l}, \mathbf{y}^{l})$ as the following form
\begin{equation}
\begin{split}
 \Big(R_{\varepsilon}(f_{\varphi}) -  \mathbb{E}_{(\mathbf{x}, \mathbf{y}) \sim \hat{\varepsilon}}&{\hat{\ell}}_{f_{\varphi}}(\mathbf{x}, \mathbf{y})\Big)\\
+ \Big(\mathbb{E}_{(\mathbf{x}, \mathbf{y}) \sim \hat{\varepsilon}}{\hat{\ell}}_{f_{\varphi}}(\mathbf{x}, \mathbf{y}) - & \frac{1}{n}\sum_{l=1}^{n}\hat{\ell}_{f_{\varphi}}(\mathbf{x}^{l}, \mathbf{y}^{l})\Big),
\label{eq:excess_risk}
\end{split}
\end{equation}
and then bound the two terms separately. The first term is called estimation difference expected for the future task \cite{maurer2009ml}. To bound the first term, we need to utilize the Lipschitz property of the margin loss and the properties of Gaussian complexity. The second part of Eq~(\ref{eq:excess_risk}) is the estimation difference between the expected empirical error of the MLMC classifier's output on a new task and the average empirical errors on the data of the past tasks. We choose to use PAC learning techniques and Gaussian contraction inequality \cite{wainwright2019book} to obtain this term's upper bound. The obtained results are shown in Theorem~\ref{thm:main_1} and Theorem~\ref{thm:main_2}, respectively.

\begin{thm}
Let $\mathcal{F}$ and $\Pi_1\mathcal{F}(\mathbf{x})$ be the same as in previous theorems. For $\rho >0$, we have
\begin{equation*}
\begin{split}
R_{\varepsilon}(f_{\varphi}) \leq & \mathbb{E}_{(\mathbf{x}, \mathbf{y}) \sim \hat{\varepsilon}}{\hat{\ell}}_{f_{\varphi}}(\mathbf{x}, \mathbf{y})\\
 + & \frac{k\sqrt{2\pi}}{\rho}\mathbb{E}_{\mu \sim \varepsilon}\mathbb{E}_{(\mathbf{x}, \mathbf{y}) \sim \mu^{m}}\Gamma(\Pi_1\mathcal{F}(\mathbf{x})).
\end{split}
\end{equation*}
\label{thm:main_1}
\end{thm}

\begin{thm}
Let $\mathcal{F}$ and $\Pi_1\mathcal{F}(\mathbf{X})$ be the same as in previous theorems. For any $\delta > 0$, we have, with probability at least $1-\delta$ on the draw of the sample $((\mathbf{x}^{1},\mathbf{y}^{1}),...,(\mathbf{x}^{n},\mathbf{y}^{n}))$,
\begin{equation*}
\begin{split}
\mathbb{E}_{(\mathbf{x}, \mathbf{y}) \sim \hat{\varepsilon}}{\hat{\ell}}_{f_{\varphi}}(\mathbf{x}, \mathbf{y}) \leq & \frac{1}{n}\sum_{l=1}^{n}\hat{\ell}_{f_{\varphi}}(\mathbf{x}^{l}, \mathbf{y}^{l})  + \sqrt{\frac{\ln(1/\delta)}{2n}}\\
+ & \frac{k\sqrt{2m\pi}}{\rho}\mathbb{E}_{(\mathbf{X},\mathbf{Y}) \sim \hat{\varepsilon}^{n}}\Gamma(\Pi_{1}\mathcal{F}(\mathbf{X})).
\end{split}
\end{equation*}
\label{thm:main_2}
\end{thm}
With these two theorems, we obtain the Gaussian complexity transfer bound for MLMC (Theorem~\ref{thm:main_gaussian_complexity}).

\subsection{Covering Number Transfer Bound for MLMC}
\label{subsect:covering_nubmer_bound}

Although we have shown a margin-based transfer bound in Theorem~\ref{thm:main_gaussian_complexity}, the value of the Gaussian complexity is still implicit. Moreover, we are more interested in the variation of $\Gamma(\Pi_1\mathcal{F}(\mathbf{X}))$ with the growth of $m$ and $n$. To this end, we need another more intuitive indicator to measure the complexity of hypothesis space $\Pi_1\mathcal{F}(\mathbf{X})$. In this work, we use the following covering number \cite{zhou2002complexity}.
\begin{defn}[Covering Number]
Let $(M,d)$ be a metric space. A subset $\hat{T}$ is called an $\epsilon$-cover of $T \subseteq M$ if $\forall t \in T$, $\exists t' \in \hat{T}$ such that $d(t, t') \leq \epsilon$. The covering number of $T$ is the cardinality of the smallest $\epsilon$-cover of $T$, that is, $\mathcal{N}(\epsilon, T,d) \triangleq \min \Big\{|\hat{T}|\ \Big|\ \hat{T}\ is\ an\ \epsilon-cover\ of\ T\ \Big\}.$
\label{def:covering_number}
\end{defn}
Let $(\mathcal{F}_{x_1,...,x_m}, \mathcal{L}_2(\widehat{D}))$ be the data-dependent $\mathcal{L}_{2}$ metric space given by metric $d(f,\hat{f})\triangleq \|f-\hat{f}\|_{2}= \sqrt{\frac{1}{m}\sum_{i=1}^{m}\big(f(x_i)-\hat{f}(x_i)\big)^{2}}$, where $\mathbf{x}=(x_1,..., x_m)$ is a sample from space $\mathcal{X}$ and $\mathcal{F}_{x_1,...,x_m}$ represents for the restriction of real-valued function class $\mathcal{F}$ to that sample. $\mathcal{N}(\epsilon, \mathcal{F}, \mathcal{L}_{2}(\mathbf{X}))$ can be defined in a similar way with the data-dependent metric $d(f,\hat{f}) = \sqrt{\frac{1}{mn}\sum_{i=1}^{n}\sum_{j=1}^{m}\big(f(x^i_j)-\hat{f}(x^i_j)\big)^{2}}$ $(f,\hat{f} \in \mathcal{F})$. Using the chaining technique \cite{talagrand2014springer}, the following refined theorem reveals the relationship between the Gaussian complexity and covering number.

\begin{thm}[Refined Dudley Entropy Bound]
For any real-valued function class $\mathcal{F}$ containing function $f: \mathcal{X} \rightarrow \mathbb{R}$, assume that $\sup_{f \in \mathcal{F}}\|f\|_{2}$ is bounded under the $\mathcal{L}_{2}(\mathbf{x})$ and $\mathcal{L}_{2}(\mathbf{X})$ metric respectively. Then
\begin{equation*}
\begin{split}
\Gamma(\mathcal{F}(\mathbf{x})) \leq \frac{24}{\sqrt{m}}\int_{0}^{\sup_{f \in \mathcal{F}}\|f\|_{2}}\sqrt{\log{\mathcal{N}(\tau, \mathcal{F}, \mathcal{L}_{2}(\mathbf{x}))}}\mathrm{d}\tau,\\
\Gamma(\mathcal{F}(\mathbf{X})) \leq \frac{24}{\sqrt{nm}}\int_{0}^{\sup_{f \in \mathcal{F}}\|f\|_{2}}\sqrt{\log{\mathcal{N}(\tau, \mathcal{F}, \mathcal{L}_{2}(\mathbf{X}))}}\mathrm{d}\tau.
\end{split}
\end{equation*}
\label{thm:dudley_entropy_bound}

\end{thm}
Bounding the Gaussian complexity in Theorem~\ref{thm:main_gaussian_complexity} with the Dudley integral in Theorem~\ref{thm:dudley_entropy_bound}, we obtain the following covering number transfer bound for MLMC.
\begin{thm}[\textbf{Margin-Based Transfer Bound for MLMC with Covering Number}]
Let $\mathcal{F}$ and $\Pi_1\mathcal{F}$ be the same as in previous theorems. Given a classification algorithm $f$ and a margin parameter $\rho >0$, let $L = \sup_{f \in \Pi_1\mathcal{F}}\|f\|_{2}$, then for any environment $\varepsilon \in M_1(M_1(\mathcal{X} \times \mathcal{Y}))$ and for any $\delta > 0$, with probability at least $1-\delta$ on the data $(\mathbf{X}, \mathbf{Y}) \sim {\hat{\varepsilon}}^{n}$, we have for all feature maps $\varphi \in \mathcal{D}$ that
\begin{equation*}
\begin{split}
& R_{\varepsilon}(f_{\varphi})\leq  \frac{1}{n}\sum_{l=1}^{n}\hat{\ell}_{f_{\varphi}}(\mathbf{x}^{l}, \mathbf{y}^{l})+ \sqrt{\frac{\ln(1/\delta)}{2n}}\\
& +   \frac{24k\sqrt{2\pi}}{\rho\sqrt{n}}\mathbb{E}_{(\mathbf{X},\mathbf{Y}) \sim \hat{\varepsilon}^{n}}\int_{0}^{L}\sqrt{\log{\mathcal{N}(\tau, \Pi_{1}\mathcal{F}, \mathcal{L}_{2}(\mathbf{X}))}}\mathrm{d}\tau\\
& +  \hspace{-0.02in} \frac{24k\sqrt{2\pi}}{\rho\sqrt{m}}\mathbb{E}_{\mu \sim \varepsilon}\mathbb{E}_{(\mathbf{x}, \mathbf{y}) \sim \mu^{m}} \hspace{-0.03in} \int_{0}^{L}\hspace{-0.05in} \sqrt{\log{\mathcal{N}(\tau, \Pi_{1}\mathcal{F}, \mathcal{L}_{2}(\mathbf{x}))}}\mathrm{d}\tau.
\end{split}
\end{equation*}
\label{thm:main_covering_number}
\end{thm}

\begin{thm}[Theorem 2.6.7 in \cite{vaart1996springer}]
Let $\mathcal{F}$ be a real-valued function class on $\mathcal{X}$ with VC-dimension $v$. Assume that $\mathcal{F}$ is uniformly bounded by $b>0$. Then, for any probability distribution $Q$ on $\mathcal{X}$,
\begin{equation*}
\mathcal{N}(\tau, \mathcal{F}, \|\cdot\|_{L_{p}(Q)}) \leq C_{0}(v+1)(16e)^{v+1}(\frac{b}{\tau})^{pv},
\end{equation*}
\label{thm:covering_number_vc_bound}
where $C_{0} >0$ is a uniform constant, and for any $f, g \in \mathcal{F}, \|f-g\|_{L_{p}(Q)}=(\int |f-g|^{p}dQ)^{1/p}, p \geq 1$.
\end{thm}

Further, since $|f(x_i, y)| \leq b$, we have $\sup_{f \in \Pi_1\mathcal{F}}\|f\|_{2} = \sup_{f \in \Pi_1\mathcal{F}} \sqrt{\frac{1}{m}\sum_{i=1}^{m}f(x_i, y)^2} \leq \sqrt{\frac{1}{m}\sum_{i=1}^{m}b^2} = b$. Combining Theorem~\ref{thm:main_covering_number} and Theorem~\ref{thm:covering_number_vc_bound}, we then obtain our most important theoretical result in Theorem~\ref{thm:transfer_bound_fsl}.

\begin{table*}[th]
\centering
\caption{The 5-way $s$-shot classification results on the \textbf{miniImageNet} dataset. We report the average accuracy (\%, top-1) as well as the 95\% confidence interval over all 600 test episodes. We compare the \textbf{Multi-Margin} loss with the \textbf{Cross-Entropy} loss. We also give the results of Baseline++ and MAML (with ${\ddagger}$), to which our current theoretical analysis is nevertheless not applicable, for comprehensive performance comparison. }
\scalebox{0.78}{
\begin{tabular}{l|cc|cc|cc}
\toprule[2pt]
     & \multicolumn{2}{|c}{\textbf{5-way 5-shot}} & \multicolumn{2}{|c}{\textbf{5-way 10-shot}} & \multicolumn{2}{|c}{\textbf{5-way 20-shot}}\\
    \textbf{Model} &\textbf{Cross-Entropy} & \textbf{Multi-Margin} & \textbf{Cross-Entropy} & \textbf{Multi-Margin} & \textbf{Cross-Entropy} & \textbf{Multi-Margin}\\
    \midrule[1pt]
    Baseline++$^{\ddagger}$\cite{yu2019iclr} & $67.00\pm0.63$ & $67.68\pm0.66$ & $73.43\pm0.59$ & $73.48\pm0.60$ & $77.32\pm0.57$ & $77.06\pm0.55$\\
    MAML$^{\ddagger}$ \cite{finn2017icml} & $61.56\pm0.70$ & $60.10\pm0.70$ & $65.76\pm0.74$ & $62.05\pm0.77$ & $65.88\pm0.77$ & $63.36\pm0.75$\\
    \midrule[1pt]
    MatchingNet \cite{oriol2016nips} & $63.68\pm0.68$ & $64.08\pm0.66$ & $68.44\pm0.66$ & $69.46\pm0.67$ & $72.64\pm0.61$ & $73.83\pm0.59$\\
    ProtoNet \cite{jake2017nips} & $65.63\pm0.73$ & $66.96\pm0.68$ & $70.28\pm0.63$ & $71.18\pm0.61$ & $74.47\pm0.75$ & $75.17\pm0.57$\\
    RelationNet \cite{sung2018cvpr} & $65.89\pm0.64$ & $65.10\pm0.67$ & $69.79\pm0.63$ & $68.73\pm0.64$ & $72.84\pm0.59$ & $71.63\pm0.58$ \\
    MetaOptNet \cite{lee2019cvpr} & $68.60\pm0.66$ & $67.96\pm0.65$ & $72.64\pm0.64$ & $71.32\pm0.59$ & $77.48\pm0.54$ & $76.83\pm0.53$\\
    \bottomrule[2pt]
  \end{tabular}
}
\label{tab:main_mini}
\end{table*}

\subsection{From Theory to Implementation}

In practical implementation, a multi-margin loss \cite{pytorch2017nips} is preferable in the multiclass classification problem, because of the convexity of the loss function \cite{mohri2012book}. The multi-margin loss on an input-output pair $(x_i, y_i)$ is $\Psi(x_{i}, y_{i})= \frac{1}{k-1}\sum_{y \neq y_{i}}^{k}\max\big(0, 1-(f_{\varphi}(x_i,y_i)- f_{\varphi}(x_i,y))/\rho\big)$. Define the empirical multi-margin loss on one training task $\tilde{\ell}_{f_{\varphi}}(\mathbf{x}^{l}, \mathbf{y}^{l}) = \frac{1}{m}\sum_{i=1}^{m}\Psi(x^{l}_{i}, y^{l}_{i})$. Due to the relationship between margin loss and multi-margin loss $\Phi(\rho_{f_{\varphi}}(x_i, y_i)) \leq (k-1)\Psi(x_i, y_i)$, we can replace the empirical margin loss $\hat{\ell}_{f_{\varphi}}(\mathbf{x}^{l}, \mathbf{y}^{l})$ in Theorem~\ref{thm:transfer_bound_fsl} with $\tilde{\ell}_{f}(\mathbf{x}^{l}, \mathbf{y}^{l})$:
\begin{equation*}
\begin{split}
R_{\varepsilon}(f_{\varphi})& \leq \frac{k-1}{n}\sum_{l=1}^{n}\tilde{\ell}_{f_{\varphi}}(\mathbf{x}^{l}, \mathbf{y}^{l}) + \sqrt{\frac{\ln(1/\delta)}{2n}} \\
 &  + (\frac{k}{\rho\sqrt{m}}+\frac{k}{\rho\sqrt{n}})(C_1 \sqrt{v} + C_2).
\label{eq:multimarginloss}
\end{split}
\end{equation*}
The multi-margin loss can be considered as the surrogate of the margin loss for easier model optimization. Next, we will use the multi-margin loss to train MLMC models.

\section{Experiments}
\label{sect:experiments}

In this section, we conduct experiments on three benchmarks to evaluate the performance of existing MLMC methods when the commonly-used cross-entropy loss is replaced by the multi-margin loss. Our main goal is to validate the practical value of our margin-based theoretical analysis for meta-learning based multiclass classification, since in some cases the margin loss or hinge loss may cause the problem of gradient vanishing in stochastic gradient descent.

\subsection{Experiment Setup}

\label{subsect:experiment setup}
\textbf{Datasets and Backbone}.
(1) \textbf{miniImageNet}. This dataset is widely used for the conventional MLMC setting, which consists of 100 classes selected from ILSVRC-2012 \cite{Russakovsky2015ImageNet}. Each class has 600 images. We use 64/16/20 classes for training/validation/test. (2) \textbf{CUB}. Under the fine-grained MLMC setting, we choose the CUB-200-2011 (CUB) dataset \cite{WahCUB_200_2011}, which has 200 classes and 11,788 images of birds. As in \cite{yu2019iclr}, we use 100/50/50 classes for training/validation/test. (3) \textbf{miniImageNet $\rightarrow$ CUB}. Under the cross-domain MLMC setting, we use 100 classes from miniImageNet for training, and 50/50 classes from CUB for validation/test, as in \cite{yu2019iclr}. For all experiments, we use a four-layer convolutional neural network (Conv-4) \cite{oriol2016nips} as the backbone with the input size of $84 \times 84$.

\noindent\textbf{Episode Sampling}.
Under the standard MLMC setting \cite{oriol2016nips, jake2017nips}, an episode is actually a training sample drawn from one task in an environment. A $k$-way $s$-shot $q$-query episode contains $k(s+q)$ images (MLMC is thus instantiated as $k$-way $s$-shot classification). We set $k=5, q=15, s=5/10/20$ during both training and test stages. Particularly, we train Baseline++ \cite{yu2019iclr} (mini-batch-training based) with 400 epochs (batch size = 256), and train meta-learning based methods with 40,000 episodes. When applied to analyze the $k$-way $s$-shot setting, our transfer bound becomes $(\sqrt{k}/(\rho\sqrt{s+q})+k/(\rho\sqrt{n}))(C_1 \sqrt{v} + C_2)$ due to $m =k(s+q)$.

\noindent\textbf{Evaluation Protocols}.
We make performance evaluation on the test set under the 5-way 5-shot, 5-way 10-shot and 5-way 20-shot settings (15-query for each test episode). Concretely, we randomly sample 600 episodes from the test set, and then report the average accuracy (\%, top-1) as well as the 95\% confidence interval over all the test episodes.

\noindent\textbf{Baselines for Comparison}.
We select six representative baselines for $k$-way $s$-shot classification: (\romannumeral 1) mini-batch-training method: Baseline++ \cite{yu2019iclr} first learns the classifier with the standard supervised training strategy, and then finetunes it to each task in the test stage. (\romannumeral 2) Meta-learning based methods: three metric-learning based methods (MatchingNet \cite{oriol2016nips}, ProtoNet \cite{jake2017nips}, RelationNet \cite{sung2018cvpr}) which just run a forward process to obtain the test accuracies during the test stage, without updating the parameters of the feature extractor; one classifier-learning based method (MetaOptNet \cite{lee2019cvpr}) which fixes the parameters of the feature extractor to extract image features and trains a new SVM classifier in each new task; one gradient based method (MAML \cite{finn2017icml}) that updates the parameters of both the feature extractor and the fully connected layer in each new task. Note that our theoretical analysis only holds for metric-learning and classifier-learning based models. For performance comparison, both the cross-entropy loss and the multi-margin loss are used for the above six representative baselines.

\noindent\textbf{Implementation Details}.
Our implementation is based on PyTorch \cite{pytorch2017nips}. We train all models from scratch and use the Adam optimizer with the initial learning rate $10^{-3}$. We select other hyperparameters (including $\rho$) by performing validation on the validation set.

\subsection{Main Results}

\begin{table*}[t]
\centering
\caption{The 5-way $s$-shot classification results on the \textbf{CUB} dataset. We report the average accuracy (\%, top-1) as well as the 95\% confidence interval over all 600 test episodes. }
\scalebox{0.78}{
\begin{tabular}{l|cc|cc|cc}
\toprule[2pt]
     & \multicolumn{2}{|c}{\textbf{5-way 5-shot}} & \multicolumn{2}{|c}{\textbf{5-way 10-shot}} & \multicolumn{2}{|c}{\textbf{5-way 20-shot}}\\
    \textbf{Model} &\textbf{Cross-Entropy} & \textbf{Multi-Margin} & \textbf{Cross-Entropy} & \textbf{Multi-Margin} & \textbf{Cross-Entropy} & \textbf{Multi-Margin}\\
    \midrule[1pt]
    Baseline++$^{\ddagger}$\cite{yu2019iclr} & $79.39\pm0.64$ & $66.24\pm0.78$ & $84.68\pm0.58$ & $78.69\pm0.63$ & $86.79\pm0.61$ & $80.31\pm0.62$ \\
    MAML$^{\ddagger}$\cite{finn2017icml} & $75.81\pm0.73$ & $74.46\pm0.70$ & $79.86\pm0.68$ & $77.12\pm0.75$ & $81.82\pm0.68$ & $78.93\pm0.78$ \\
    \midrule[1pt]
     MatchingNet \cite{oriol2016nips} & $78.93\pm0.64$ & $79.20\pm0.62$ & $83.72\pm0.56$ & $84.33\pm0.56$ & $86.93\pm0.45$ & $87.03\pm0.47$ \\
   ProtoNet \cite{jake2017nips} & $79.03\pm0.67$ & $79.88\pm0.68$ & $83.84\pm0.59$ & $84.62\pm0.59$ & $86.85\pm0.52$ & $87.39\pm0.55$ \\
    RelationNet \cite{sung2018cvpr} & $79.18\pm0.65$ & $77.84\pm0.68$ & $81.34\pm0.57$ & $80.28\pm0.61$ & $82.17\pm0.54$ & $81.46\pm0.59$ \\
    MetaOptNet \cite{lee2019cvpr} & $79.23\pm0.71$ & $78.08\pm0.67$ & $83.92\pm0.60$ & $82.25\pm0.64$ & $87.12\pm0.57$ & $86.78\pm0.61$ \\
    \bottomrule[2pt]
  \end{tabular}
}
\label{tab:main_cub}
\end{table*}

\begin{figure*}[th]
\vspace{0.2in}
\centering
\subfigure[]{
\includegraphics[height=0.31\textwidth]{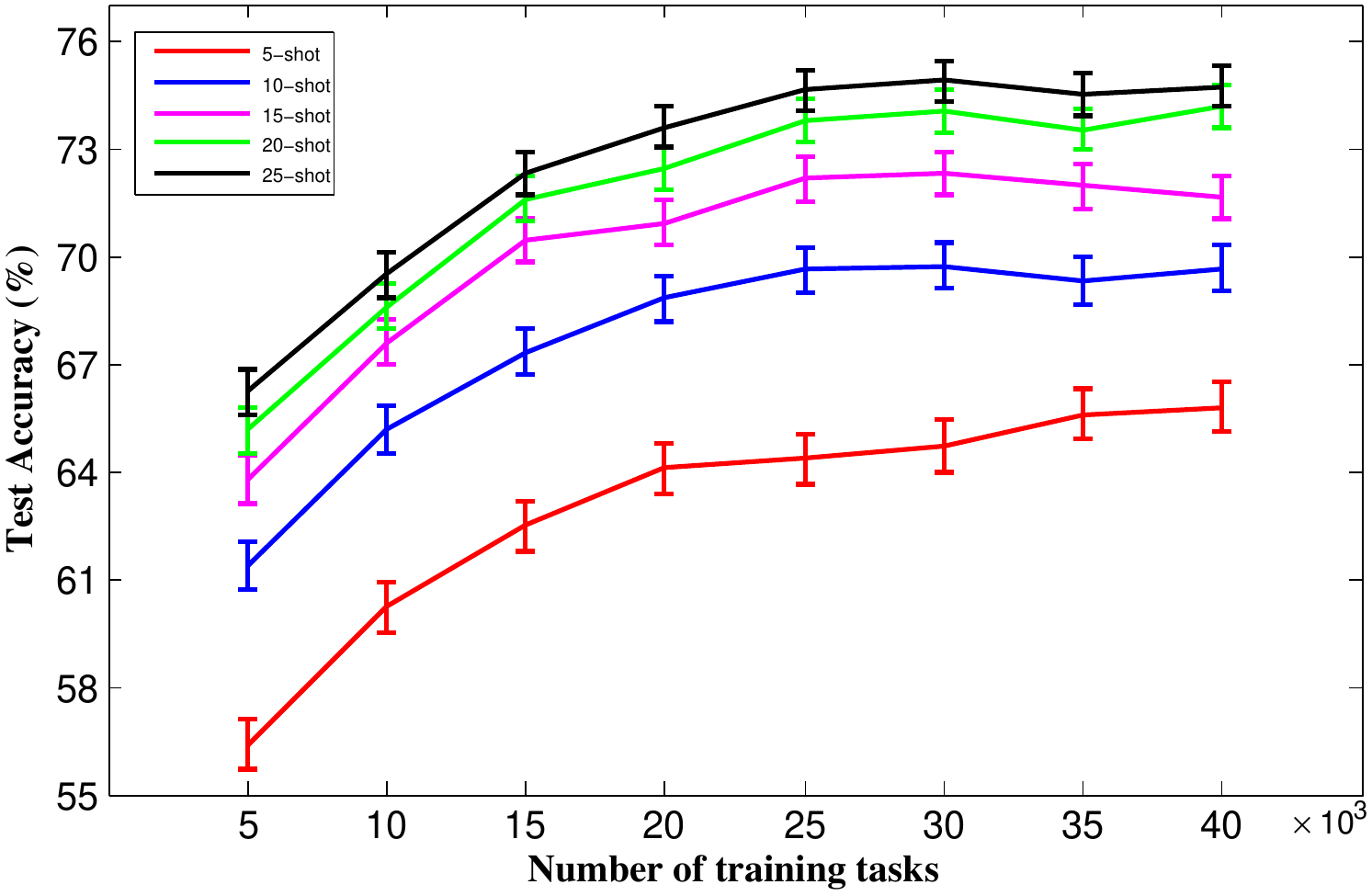}
\label{fig:number_training_episodes}
}
\subfigure[]{
\includegraphics[height=0.31\textwidth]{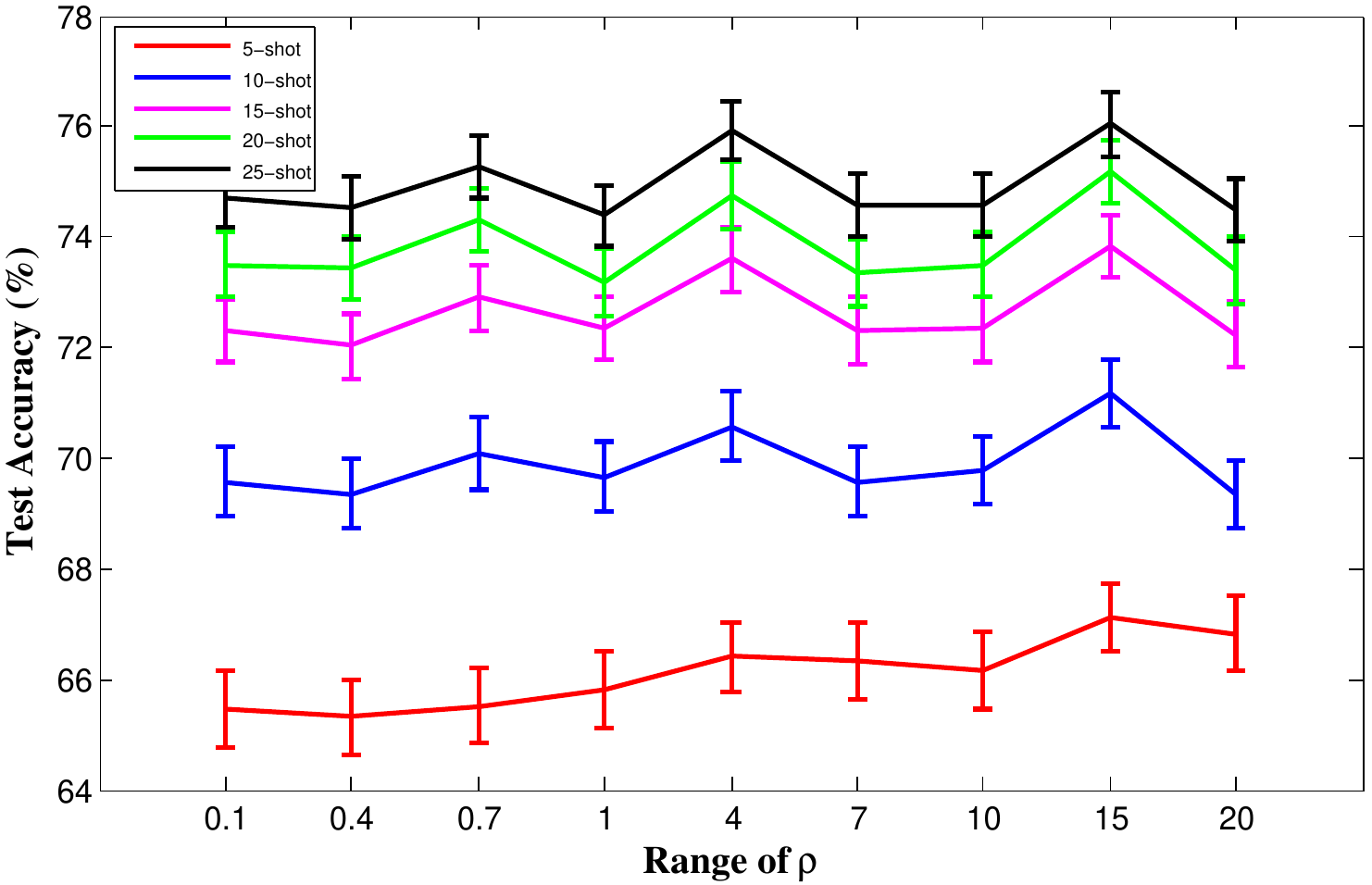}
\label{fig:range_rho}
}
\vspace{-0.1in}
\caption{(a) The 5-way test accuracies of ProtoNet with different numbers of training tasks on miniImageNet ($\rho = 1$). (b) The 5-way test accuracies of ProtoNet with different choices of $\rho$ on miniImageNet ($n=$40,000). For both figures, different choices of $s$-shot are also considered. }
\label{fig:influence_hyperparameter}
\end{figure*}

\vspace{0.1in}
\begin{table}
\centering
\caption{Comparative results on the cross-domain \textbf{miniImageNet $\rightarrow$ CUB} dataset. Average 5-way 5-shot classification accuracies (\%) with 95\% confidence intervals are computed here. More $s$-shot classification results can be found in the supplementary material.}
\scalebox{0.7}{
\begin{tabular}{lcc}
\toprule[2pt]
    \textbf{Model} & \textbf{Cross-Entropy} & \textbf{Multi-Margin}\\
    \midrule[1pt]
    Baseline++ \cite{yu2019iclr} & $66.21\pm0.70$ & $67.90\pm0.69$\\
    MAML \cite{finn2017icml} & $61.09\pm0.57$ & $60.85\pm0.54$\\
    \midrule[1pt]
     MatchingNet \cite{oriol2016nips} & $64.47\pm0.73$ & $63.34\pm0.71$ \\
    ProtoNet \cite{jake2017nips} & $64.51\pm0.75$  & $64.41\pm0.75$ \\
    RelationNet \cite{sung2018cvpr} & $64.68\pm0.74$ & $65.16\pm0.74$  \\
    MetaOptNet \cite{lee2019cvpr} & $65.17\pm0.73$ & $64.62\pm0.74$  \\
    \bottomrule[2pt]
  \end{tabular}
}
\label{tab:cross_domain_fsl}
\end{table}

\noindent\textbf{Performance of Multi-Margin Loss}.
We focus on comparing the multi-margin loss with the cross-entropy loss when both are used for $k$-way $s$-shot classification. From Tables~\ref{tab:main_mini}--\ref{tab:cross_domain_fsl}, we have the following observations: (\romannumeral 1) In most cases, the classification performance obtained with the multi-margin loss is comparable to that obtained with the cross-entropy loss. (\romannumeral 2) On the CUB dataset, the inferior performance of Baseline++ with the multi-margin loss suggests that the margin loss may be unsuitable for mini-batch-training model optimization in fine-grained classification. (\romannumeral 3) For meta-learning based classification methods, the multi-margin loss generally leads to competitive results (w.r.t. the the cross-entropy loss), which is consistent with our margin-based theoretical analysis.

\noindent\textbf{Influence of Hyperparameters}.
We further conduct experiments to study the influence of three hyperparameters -- the number of the training tasks $n$, the sample size per episode $m~(=ks)$ and the margin parameter $\rho$ -- on the performance of meta-learning based classification. We select ProtoNet as the baseline and show the 5-way test accuracies (i.e. $k=5$) on miniImageNet in Figure~\ref{fig:influence_hyperparameter}. We can find that: (\romannumeral 1) Meta-learning based classification with the multi-margin loss can achieve higher test accuracies with the growth of $n$ and $m$, but the improvements are incremental when $n$ or $m$ is large (e.g. $n\geq 30,000, m\geq 100$ (i.e. $s\geq 20$)). (\romannumeral 2) The performance of meta-learning based classification with the multi-margin loss is not so sensitive to the variations of $\rho$.

\section{Related Work}
\label{sect:related_work}

The pioneering work \cite{baxter2000jai} on meta-learning theory differs from our results in several aspects: (1) \cite{baxter2000jai} is not focused on the multiclass classification problem and does not explicitly reveal the relationship between the transfer bound and the number $k$ of classification categories, but we show that our transfer bounds admit only a linear dependency on $k$. (2) Its transfer bound with a neural network feature map (Theorem 8 in \cite{baxter2000jai}) depends on the feature dimension, while our transfer bounds (e.g. Theorem~\ref{thm:transfer_bound_fsl}) are dimension free. (3) The theoretical results about feature maps in \cite{baxter2000jai} pay more attention to simple two-layer neural networks and mainly use the neural network for feature dimension reduction. However, this paper focuses on the role of deep feature embedding in image feature extraction, which is the modern development of feature engineering in machine learning community.

Following \cite{baxter2000jai}, recent theoretical works can be divided into two main groups: one group explores PAC-Bayes theory \cite{pentina2014icml, dziugaite2017uai, amit2018icml} for meta learning, and the other utilizes conventional PAC learning techniques (i.e. without prior assumptions) to give transfer bounds for different models (e.g. regression learner \cite{maurer2009ml, maurer2016jmlr, denevi2018nips} or algorithm stability \cite{maurer2005jmlr}). Below, we detail our differences from the two most related works \cite{amit2018icml, maurer2009ml}.\\

\noindent\textbf{PAC-Bayes Meta Learning Theory}. It assumes a prior distribution over priors, a `hyper-prior' $\mathcal{P}(P)$, and after training outputs a distribution over priors, a `hyper-posterior' $\mathcal{Q}(P)$. Let $er(\mathcal{Q}, \varepsilon)= \mathbb{E}_{P \sim \mathcal{Q}}er(P, \varepsilon)$ be the expected loss to measure the quality of the prior $Q$, where $er(P, \varepsilon)\triangleq \mathbb{E}_{\mu \sim \varepsilon}\mathbb{E}_{S \sim \mu^m}\mathbb{E}_{f \sim Q(S, P)}\mathbb{E}_{(x,y) \sim \mu}l(f, x, y)$ ($l$ is the loss function). $S_i=(\mathbf{x}^{i},\mathbf{y}^{i})$ is a sample and $er(Q, S_i)=\mathbb{E}_{f \sim Q}\frac{1}{m}\sum_{j=1}^{m}l(f, x_{j}^{i},y_{j}^{i})$ is the empirical loss. $ D(Q||P)=\mathbb{E}_{f\sim Q}\ln\frac{Q(f)}{P(f)}$ represents the Kullback-Leibler Divergence between two distributions $Q$ and $P$.
\begin{thm}[\textbf{Theorem 2 in} \cite{amit2018icml}]
Let $Q: S \times \mathcal{F} \rightarrow \mathcal{F}$ be a base-learner and $Q_i\triangleq Q(S_i, P)$. For any hyper-posterior $\mathcal{Q}$ and for any $\delta>0$, with probability as least $1-\delta$, the inequality holds:
\begin{small}
\begin{equation*}
\begin{split}
er(\mathcal{Q}, \varepsilon)\hspace{-0.04in} \leq &  \frac{1}{n}\hspace{-0.02in}\sum_{i=1}^{n}\hspace{-0.02in}\hspace{-0.02in}\mathop{\mathbb{E}}\limits_{P \sim \mathcal{Q}}\hspace{-0.04in}\hat{er}_{i}(Q_i,S_i)\hspace{-0.02in}+ \hspace{-0.02in}+ \hspace{-0.02in} \frac{\sqrt{{D(\mathcal{Q}||\mathcal{P})+\log\frac{2n}{\delta}}}}{\sqrt{{2n-2}}}\\
+ &\frac{1}{n}\hspace{-0.02in}\sum_{i=1}^{n} \hspace{-0.02in}\frac{\sqrt{D(\mathcal{Q}||\mathcal{P})+\mathop{\mathbb{E}}\limits_{P\sim \mathcal{Q}}D(Q_{i}||P)+\log\frac{2nm}{\delta}}}{{\sqrt{2m-2}}}.
\end{split}
\end{equation*}
\end{small}
\label{thm:meta_learning_pac_bayes}
\end{thm}
The transfer bound in Theorem~\ref{thm:meta_learning_pac_bayes} differs from ours in Theorem~\ref{thm:transfer_bound_fsl} in two aspects: (\romannumeral 1) PAC-Bayes bound is expressed on the averaging of multiple hypothesises (weighted by a posterior distribution), but our bound is expressed for any hypothesis. (\romannumeral 2) Even ignoring the KL-divergence term between two distinct distributions, the PAC-Bayes bound has still a complexity part $\mathcal{O}(\frac{\sqrt{\log(nm)}}{\sqrt{m}}+\frac{\sqrt{\log n}}{\sqrt{n}})$, while we derive a  bound $\mathcal{O}(k(\sqrt{\frac{v}{m}} + \sqrt{\frac{v}{n}}))$ which is generally lower.\\

\noindent\textbf{Transfer Bounds for Linear Regression}.
Another closely related work is \cite{maurer2009ml} which focuses on linear regression problems. Below, we show the transfer bound of the regression learners in \cite{maurer2009ml}, and compare it with our Theorem~\ref{thm:transfer_bound_fsl}. Considering the regularized least squares regression, we have the weight vectors $\omega(\mathbf{x},\mathbf{y}) = \arg\min_{\omega \in H}\big( \frac{1}{m}\sum_{i=1}^{m}(\left<\omega, x_{i}\right>-y_{i})^{2} + \|\omega\|^{2}\big)$. The empirical error is $\hat{\ell}_{\omega}(\mathbf{x}, \mathbf{y})= \frac{1}{m}\sum_{i=1}^{m}(\left<\omega(\mathbf{x},\mathbf{y}), x_{i}\right> - y_{i})^{2}$. $\mathcal{P}_{d}$ is the set of orthogonal projections $P$ with $d$-dimension, and $\hat{\ell}_{\omega_{\lambda^{-1}P}}(\mathbf{x}^{l}, \mathbf{y}^{l}) = \hat{\ell}_{\omega}(\lambda^{-1/2}P^{1/2}\mathbf{x}^{l}, \mathbf{y}^{l})$ ($\lambda > 0$ is the regularization parameter). Let $\|C\|_{\infty}$ be the largest eigenvalue of the covariance operator $C$ for the total input distribution.
\begin{thm}[\textbf{Theorem 1 in} \cite{maurer2009ml}]
For any $\delta > 0$, we have with probability at least $1-\delta$ on the data $(\mathbf{X}, \mathbf{Y}) = ((\mathbf{x}^{1}, \mathbf{y}^{1}), ..., (\mathbf{x}^{n}, \mathbf{y}^{n}))$ that for all feature maps $P \in \mathcal{P}_{d}$
\begin{equation*}
\begin{split}
R_{\varepsilon}(\omega_{\lambda^{-1}P}) \leq & \frac{1}{n}\sum_{l=1}^{n}\hat{\ell}_{\omega_{\lambda^{-1}P}}(\mathbf{x}^{l}, \mathbf{y}^{l}) + \sqrt{\frac{\ln(1/\delta)}{2n}} \\
+ & \frac{\sqrt{8\pi d}}{\lambda}\big(2\sqrt{\frac{\|C\|_{\infty}}{m}}+\sqrt{\frac{1}{n}}\big).
\end{split}
\end{equation*}
\label{thm:linear_regression}
\end{thm}

Note that the transfer bounds for regression and classification problems have a similar form $\mathcal{O} (\frac{1}{\sqrt{m}} + \frac{1}{\sqrt{n}})$, if we ignore the dependence on the confidence parameter $\delta$ in Theorem~\ref{thm:linear_regression} \& Theorem~\ref{thm:transfer_bound_fsl} and assume that the VC-dimension $v$ in Theorem~\ref{thm:transfer_bound_fsl} is finite. That is, although our work is quite different from \cite{maurer2009ml} (i.e. nonlinear classification vs. linear regression), our transfer bounds are somewhat similar to the transfer bound of \cite{maurer2009ml}, indirectly showing the correct derivation of our transfer bounds.

Finally, we stress that the sample efficiency per task can also be guaranteed by our transfer bounds for meta learning with deep feature embedding. Specifically, given the accuracy $\epsilon$, to let the inequality $R_{\varepsilon}(f_{\varphi})\leq \frac{1}{n}\sum_{l=1}^{n} \hat{\ell}_{f_{\varphi}}(\mathbf{x}^{l}, \mathbf{y}^{l}) + \epsilon$ hold, which means our transfer bound $\mathcal{O}(k(\sqrt{\frac{v}{m}} + \sqrt{\frac{v}{n}}))= ak(\sqrt{\frac{v}{m}} + \sqrt{\frac{v}{n}}) \leq \epsilon$ ($a$ is a constant), we must let the number of examples per task required for good generalization obey $m \geq \frac{a^2k^2v}{{(\epsilon -ak\sqrt{\frac{v}{n}})}^2}$ $(\triangleq \phi(n))$. Since $\phi(n)$ is a monotonically decreasing function (w.r.t. $n$), the number $m$ of examples per task required for good generalization will decrease as the number $n$ of tasks increases. Therefore, our theoretical results guarantee the sample efficiency per task in meta learning. Similarly, such sample efficiency per task can also be guaranteed by \cite{maurer2009ml}, though it focuses on the regression problem instead.

\section{Conclusion and Future Work}
\label{sect:conclusions}

We have derived margin-based transfer bounds for meta-learning based multiclass classification, showing that its expected error on a future task can be properly estimated by its empirical error on previous tasks. We show that our transfer bounds only admit a linear dependency on classification categories, and point out the importance of the choice of deep feature embedding in meta-learning. The experiment results demonstrate the practical significance of our margin-based theoretical analysis. Our ongoing research includes: (\textbf{\romannumeral 1})~The cross-entropy loss is the most common choice for multiclass classification, and performs better than the multi-margin loss in some cases. One research direction is to explore whether we can use the cross-entropy loss to obtain similar theoretical results. (\textbf{\romannumeral 2}) Our most important theoretical result is given by using Gaussian complexity and Slepian's Lemma in Gaussian process. Is it possible to obtain a similar or tighter transfer bound via more concise theoretical analysis?


\bibliography{mlmc_bound}

\newpage
\begin{center}
\MakeUppercase{\textbf{\large{Supplementary Material}}}
\end{center}
In the supplementary material, Section~\nameref{suppl_sect:auxiliary_results} provides auxiliary results which help prove our main theorems for meta-learning based multiclass classification (MLMC). Section~\nameref{suppl_sect:gaussian_complexity_bound} and Section~\nameref{suppl_sect:covering_number_bound} show our technical proofs for transfer bounds with Gaussian complexity and covering number, respectively. Section~\nameref{suppl_sect:transfer_bound_fsl} gives the demonstrations for our most important result, the margin-based transfer bound with VC-dimension (i.e. Theorem~\ref{thm:transfer_bound_fsl} in main paper). Section~\nameref{suppl_sect:cross-domain mlmc} gives the comprehensive experiment results on the cross-domain dataset \textbf{miniImageNet $\rightarrow$ CUB}.

\section{Auxiliary Results}
\label{suppl_sect:auxiliary_results}

We use $\{\sigma_i: i \in \mathbb{N}\}$ to denote a sequence of independent Bernoulli variables (i.e. $\sigma_i \in \{-1, +1\}$) and use $\{\gamma_i: i\in \mathbb{N}\}$ as a sequence of independent standard Gaussian variables (i.e. $\gamma_i \sim N(0,1)$), which are also independent of $\{\sigma_i\}$. For $A \subseteq \mathbb{R}^{m}$ we define the Rademacher complexity and Gaussian complexity of $A$ as
\begin{equation*}
\mathcal{R}(A)= \mathbb{E}_{\sigma}\sup_{\mathbf{x} \in A}\frac{2}{m}\sum_{i=1}^{m}\sigma_i x_i, \quad \Gamma(A)= \mathbb{E}_{\gamma}\sup_{\mathbf{x} \in A}\frac{2}{m}\sum_{i=1}^{m}\gamma_i x_i.
\end{equation*}
Jensen's inequality implies the relationship between these two kinds of complexities
\begin{equation}
\begin{split}
\Gamma(A)= & \mathbb{E}\sup_{\mathbf{x} \in A}\frac{2}{m}\sum_{i=1}^{m}\gamma_i x_i = \mathbb{E}\sup_{\mathbf{x} \in A}\frac{2}{m}\sum_{i=1}^{m}|\gamma_i|\sigma_i x_i \\
\geq & \mathbb{E}\sup_{\mathbf{x} \in A}\frac{2}{m}\sum_{i=1}^{m}\mathbb{E}|\gamma_i|\sigma_i x_i =\sqrt{\frac{2}{\pi}}\mathcal{R}(A).
\end{split}
\label{suppl_eq:rademacher_gaussian}
\end{equation}
The following Theorem is fundamental to derive our results (e.g. Theorem~\ref{suppl_thm:main}, \ref{suppl_thm:main_1} and \ref{suppl_thm:main_2}). For the readers' benefit we present them with a sketch of proof and the detailed proof can be found in \cite{vaart1996springer} for (\romannumeral 1) and \cite{koltchinskii2002annual_stat} for (\romannumeral 2).
\begin{thm}
Let $\mathcal{F}$ be a real-valued function class on a space $\mathcal{X}$ and $\mu \in M_1(\mathcal{X})$. For $\mathbf{x}=(x_1,...,x_m) \in \mathcal{X}^{m}$ define
\begin{equation*}
\Phi(\mathbf{x}) = \sup_{f \in \mathcal{F}}\bigg(\mathbb{E}_{x \sim \mu}[f(x)]- \frac{1}{m}\sum_{i=1}^{m}f(x_i)\bigg).
\end{equation*}
\rm{(\romannumeral 1)}\quad $\mathbb{E}_{\mathbf{x}\sim\mu^{m}}[\Phi(\mathbf{x})] \leq \mathbb{E}_{\mathbf{x}\sim\mu^{m}} \mathcal{R}(\mathcal{F}(\mathbf{x}))$.\\
\rm{(\romannumeral 2)}\quad If $\mathcal{F}$ is [0,1]-valued then $\forall \delta > 0$ we have with probability greater than $1-\delta$ in $\mathbf{x} \sim \mu^{m}$ that
\begin{equation*}
\Phi(\mathbf{x}) \leq \mathbb{E}_{\mathbf{x}\sim\mu^{m}} \mathcal{R}(\mathcal{F}(\mathbf{x})) + \sqrt{\frac{\ln(1/\delta)}{2m}}.
\end{equation*}
\rm{(\romannumeral 3)}\quad $\mathcal{R}(\mathcal{F}(\mathbf{x}))$ can be replaced by $\sqrt{\pi/2}\Gamma(\mathcal{F}(\mathbf{x}))$ in \rm{(\romannumeral 1)} and \rm{(\romannumeral 2)}.
\label{suppl_thm:wellner_bartlett}
\end{thm}
\noindent\emph{Proof.} For any Rademacher variables $\sigma = \{\sigma_{i}\}_{i=1}^{m}$
\begin{equation*}
\begin{split}
\mathbb{E}_{\mathbf{x}\sim\mu^{m}}[\Phi(\mathbf{x})] = & \mathbb{E}_{\mathbf{x}\sim\mu^{m}} \sup_{f \in \mathcal{F}}\frac{1}{m}\mathbb{E}_{\mathbf{x'}\sim\mu^{m}} \sum_{i=1}^{m}\big(f(x'_{i})-f(x_{i})\big)\\
\leq & \mathbb{E}_{\mathbf{x,x'}\sim\mu^{m} \times \mu^{m}} \sup_{f \in \mathcal{F}}\frac{1}{m}\sum_{i=1}^{m}\sigma_{i}\big(f(x'_{i})-f(x_{i})\big)\\
\end{split}
\end{equation*}
The last inequality holds due to the symmetry of the measure $\mu^{m} \times \mu^{m}$ and the interchangeability between $x_{i}$ and $x'_{i}$. Taking the expectation of $\sigma$ and using the triangle inequality we obtain \rm{(\romannumeral 1)}. Then, applying the McDiarmid concentration inequality to $\Phi(\mathbf{x})$ we have with probability at least $1-\delta$, $\Phi(\mathbf{x}) \leq\mathbb{E}_{\mathbf{x}\sim\mu^{m}}[\Phi(\mathbf{x})] + \sqrt{\frac{\ln(1/\delta)}{2m}}$, and recall (\romannumeral 1) we have (\romannumeral 2). Finally, Eq.~(\ref{suppl_eq:rademacher_gaussian}) gives (\romannumeral 3). $\hfill\blacksquare$

The following theorems (Theorem~\ref{suppl_thm:slepian},~\ref{suppl_thm:gaussian_contraction_ineq} and \ref{suppl_thm:max_f}) about Gaussian complexity or Gaussian Process are needed to obtain our results (like Theorem~\ref{suppl_thm:main_1} and \ref{suppl_thm:main_2}).
\begin{thm}[Slepian's Lemma \cite{Ledoux1991springer}]
Let $\Omega$ and $\Xi$ be mean zero, separable Gaussian processes indexed by a common set $T$, such that
\begin{equation*}
\mathbb{E}(\Omega_s - \Omega_t)^{2} \leq \mathbb{E}(\Xi_s - \Xi_t)^2 \quad \forall s,t \in T.
\end{equation*}
Then $\mathbb{E}\sup_{t \in T}\Omega_t \leq \mathbb{E}\sup_{t \in T} \Xi_t$.
\label{suppl_thm:slepian}
\end{thm}

\begin{thm}[Gaussian Contraction Inequality, Exercise 5.12 in \cite{wainwright2019book}]
 Consider a bounded subset  $T \subseteq \mathbb{R}^{m}$, and let $\{\gamma_i\}_{i \geq 1}$ be independent $N(0,1)$ random variables. Let $\Phi_{i}:\mathbb{R}\rightarrow\mathbb{R}$ be $\ell$-Lipschitz contractions, i.e., $\forall x,y \in \mathbb{R}, |\Phi_{i}(x)-\Phi_{i}(y)|\leq \ell|x-y|$. Then we have
\begin{equation*}
\mathbb{E}\sup_{t \in T}\sum_{i=1}^{m}\gamma_i\Phi_{i}(t_i) \leq \ell\mathbb{E}\sup_{t \in T}\sum_{i=1}^{m}\gamma_{i}t_{i}.
\end{equation*}
\label{suppl_thm:gaussian_contraction_ineq}
\end{thm}
\noindent\emph{Proof.} Define Gaussian processes $\{\Omega_t\}_{t \in T}, \{\Xi_t\}_{t \in T}$, where $\Omega_{t} = \sum_{i=1}^{m}\gamma_{i}\Phi_{i}(t_{i})= \langle\bm{\gamma}, \bm{\Phi}(t)\rangle, \bm{\Phi}(t)=(\Phi_{1}(t_1),...,\Phi_{m}(t_{m}))^{T}$ and $\Xi_{t}=\sum_{i=1}^{m}\ell\gamma_{i}t_{i}$. We have
\begin{equation*}
\begin{split}
& \mathbb{E}(\Omega_{s} - \Omega_{t})^{2}= \mathbb{E}(\langle\bm{\gamma}, \bm{\Phi}(s)-\bm{\Phi}(t)\rangle)^{2}\\
=& \mathbb{E}\big(\bm{\Phi}(s)-\bm{\Phi}(t)\big)^{T}\bm{\gamma}\bm{\gamma}^{T}\big(\bm{\Phi}(s)-\bm{\Phi}(t)\big)\\
= & \big(\bm{\Phi}(s)-\bm{\Phi}(t)\big)^{T}\big(\bm{\Phi}(s)-\bm{\Phi}(t)\big)\qquad \rm{(\mathbb{E}\big[\bm{\gamma}\bm{\gamma}^{T}\big]= \mathbf{I})}\\
= & \sum_{i=1}^{m}\big(\Phi_{i}(s_i)-\Phi_{i}(t_i)\big)^{2}\\
\leq & \sum_{i=1}^{m}\ell^{2}(s_{i}-t_{i})^{2} \qquad \rm{(Lipschitz\ Property)}\\
= & \mathbb{E}(\Xi_{s} - \Xi_{t})^{2}.
\end{split}
\end{equation*}
From Theorem~\ref{suppl_thm:slepian} we have $\mathbb{E}\sup_{t \in T}\Omega_{t} \leq \mathbb{E}\sup_{t \in T}\Xi_{t}$. $\hfill\blacksquare$\\
Recall that $\Gamma(\mathcal{F}(\mathbf{x}))=2/m\mathbb{E}\sup_{f \in \mathcal{F}}\sum_{i=1}^{m}\gamma_{i}f(x_{i})$. If $\ell$-Lipschitz contractions $\Phi_{i}=\Phi$ for all $i \in [m]$, let $\Phi {\scalebox{.8}{$\circ$}} \mathcal{F}(\mathbf{x})=\{\big(\Phi {\scalebox{.8}{$\circ$}}f(x_1),..., \Phi {\scalebox{.8}{$\circ$}}f(x_m)\big):f \in \mathcal{F}\})$, then from Theorem~\ref{suppl_thm:gaussian_contraction_ineq}, we have
\begin{equation}
\Gamma(\Phi {\scalebox{.8}{$\circ$}} \mathcal{F}(\mathbf{x})) \leq \ell \Gamma(\mathcal{F}(\mathbf{x})).
\label{suppl_eq:talagrand}
\end{equation}
\begin{thm}
Let $\mathcal{F}_1,...,\mathcal{F}_l$ be l hypothesis sets in $\mathbb{R}^{\mathcal{X}}, l \geq 1$, and let $\mathcal{G}=\{max\{f_1,...,f_l\}:f_i \in \mathcal{F}_i, i \in [l]\}$. Then for any sample $\mathbf{x}$ of size m, we have
\begin{equation*}
\Gamma(\mathcal{G}(\mathbf{x})) \leq \sum_{i=1}^{l}\Gamma(\mathcal{F}_i(\mathbf{x})).
\end{equation*}
\label{suppl_thm:max_f}
\end{thm}
\noindent\emph{Proof.} The main idea is to notice that $max\{f_1,f_2\}=(f_{1}+f_{2}+|f_{1}-f_{2}|)/2$ and use the sub-additivity of $\sup$ function. The proof is similar to that for Lemma 9.1 in \cite{mohri2012book} (Rademacher complexity version results). The only difference is that for the Gaussian complexity, we use Gaussian Contraction Inequality which is proved in Theorem~\ref{suppl_thm:gaussian_contraction_ineq}. The detailed demonstration is left to readers. $\hfill\blacksquare$\\
To demonstrate the refined Dudley entropy bound \cite{talagrand2014springer} of Gaussian complexity (see Theorem~\ref{suppl_thm:dudley_entropy_bound} in Sect.~\ref{suppl_sect:covering_number_bound}), we need the following refined Massart lemma which focuses on bounding the finite set's Gaussian complexity with the set's cardinality.
\begin{lemma}[Refined Massart Lemma]
Let $A= \{\mathbf{a}_{1},...,\mathbf{a}_{N}\}$ be a finite set of vectors in $\mathbb{R}^{m}$. Define $\bar{\mathbf{a}} = \frac{1}{N}\sum_{i=1}^{N}\mathbf{a}_{i}$, we have
\begin{equation*}
\Gamma(A)\leq \max_{\mathbf{a} \in A}\|\mathbf{a}-\bar{\mathbf{a}}\|\frac{2\sqrt{2 \log{N}}}{m}
\end{equation*}
\label{suppl_lemma:massart}
\end{lemma}
\noindent\emph{Proof.} Without loss of generality, we assume that $\bar{\mathbf{a}}=0$. $\forall \lambda > 0$, let $A' = \{\lambda \mathbf{a}_{1},...,\lambda\mathbf{a}_{N}\}$, then
\begin{equation}
\begin{split}
& \frac{m}{2}\Gamma(A')= \mathbb{E}_{\bm{\gamma}}\sup_{\mathbf{a} \in A'}\langle\mathbf{a}, \bm{\gamma}\rangle\\
= & \mathbb{E}_{\bm{\gamma}}\log \max_{\mathbf{a} \in A'} e^{\langle\mathbf{a}, \bm{\gamma}\rangle}\\
\leq & \log \mathbb{E}_{\bm{\gamma}}\max_{\mathbf{a} \in A'}e^{\langle\mathbf{a}, \bm{\gamma}\rangle} \qquad \rm{(Jensen)}\\
\leq & \log \mathbb{E}_{\bm{\gamma}}\sum_{\mathbf{a} \in A'}e^{\langle\mathbf{a}, \bm{\gamma}\rangle}\\
= & \log \sum_{\mathbf{a} \in A'}\prod_{i=1}^{m}\mathbb{E}_{\gamma_{i}}e^{a_{i} \gamma_{i}}\\
= & \log \sum_{\mathbf{a} \in A'}\prod_{i=1}^{m}e^{\frac{a_{i}^{2}}{2}} \qquad \rm{(\int_{\mathbb{R}}\frac{1}{\sqrt{2\pi}}e^{-\frac{x^2}{2}}e^{ax}\mathrm{d}x = e^{\frac{a^2}{2}})}\\
= & \log \sum_{\mathbf{a} \in A'}e^{\frac{\|\mathbf{a}\|^{2}}{2}}\\
\leq & \log \big(N \max_{\mathbf{a} \in A'}e^{\frac{\|\mathbf{a}\|^{2}}{2}}\big)\\
= & \log N + \max_{\mathbf{a} \in A'}\frac{\|\mathbf{a}\|^{2}}{2}.\\
\end{split}
\label{suppl_eq:massart}
\end{equation}
Let $L = \max_{\mathbf{a} \in A}\|\mathbf{a}\|$, then we have $\forall \lambda > 0$
\begin{equation*}
\begin{split}
\Gamma(A)\leq & \frac{\Gamma(A')}{\lambda} \\
\leq & \frac{\frac{2\log{N}}{\lambda} + \lambda L^2}{m} \qquad \rm{by\ Eq.~(\ref{suppl_eq:massart})}\\
\end{split}
\end{equation*}
Plugging $\lambda = \sqrt{2\log N}/L$ into the above the inequality, we have $\Gamma(A)\leq \frac{2L\sqrt{2\log N}}{m}$.
$\hfill\blacksquare$

\section{Gaussian Complexity Transfer Bound}
\label{suppl_sect:gaussian_complexity_bound}

\begin{thm}[\textbf{Margin-based Transfer Bound for MLMC with Gaussian Complexity, Theorem~\ref{thm:main_gaussian_complexity} in main paper}]
Let $\mathcal{F}$ be a hypothesis of scoring functions. Given a classification algorithm $f$ and a margin parameter $\rho >0$, for any environment $\varepsilon \in M_1(M_1(\mathcal{X} \times \mathcal{Y}))$ and for any $\delta > 0$, with probability at least $1-\delta$ on the data $(\mathbf{X}, \mathbf{Y}) \sim {\hat{\varepsilon}}^{n}$, we have for all feature maps $\varphi \in \mathcal{D}$ that
\begin{equation*}
\begin{split}
R_{\varepsilon}(f_{\varphi})\leq & \hspace{-0.01in}\frac{1}{n}\sum_{l=1}^{n}\hspace{-0.01in}\hat{\ell}_{f_{\varphi}}(\mathbf{x}^{l}, \mathbf{y}^{l}) +  \frac{k\sqrt{2m\pi}}{\rho}\hspace{-0.1in}\mathop{\mathbb{E}}\limits_{(\mathbf{X},\mathbf{Y}) \sim \hat{\varepsilon}^{n}}\hspace{-0.1in}\Gamma(\Pi_{1}\mathcal{F}(\mathbf{X}))  \\
 + & \sqrt{\frac{\ln(1/\delta)}{2n}} + \frac{k\sqrt{2\pi}}{\rho}\hspace{-0.02in}\mathop{\mathbb{E}}\limits_{\mu \sim \varepsilon}\mathop{\mathbb{E}}\limits_{(\mathbf{x}, \mathbf{y}) \sim \mu^{m}}\hspace{-0.1in}\Gamma(\Pi_1\mathcal{F}(\mathbf{x})),
\end{split}
\end{equation*}
where $\Pi_1\mathcal{F}(\mathbf{X})=\{\big(f_{\varphi_{(\mathbf{X},\mathbf{Y})}}(x_{1}^{1},y),...,f_{\varphi_{(\mathbf{X},\mathbf{Y})}}(x_{m}^{1},y),\\
...,f_{\varphi_{(\mathbf{X},\mathbf{Y})}}(x_{1}^{n},y),...,f_{\varphi_{(\mathbf{X},\mathbf{Y})}}(x_{m}^{n},y)\big): y \in \mathcal{Y}, \varphi \in \mathcal{D}\}$, $\Pi_1\mathcal{F}(\mathbf{x})=\{\big(f_{\varphi(\mathbf{x},\mathbf{y})}(x_{1},y),...,f_{\varphi(\mathbf{x},\mathbf{y})}(x_{m},y)\big): y \in \mathcal{Y}, \varphi \in \mathcal{D}\}$, and the scoring function $f_{\varphi_{(\mathbf{X},\mathbf{Y})}}$ is defined as: $f_{\varphi_{(\mathbf{X},\mathbf{Y})}}(x_{i}^{l}, y) = f_{\varphi(\mathbf{x}^{l},\mathbf{y}^{l})}(x_{i}^{l},y), \forall i \in [m], l \in [n]$.
\label{suppl_thm:main}
\end{thm}
The proof strategy is to rewrite $R_{\varepsilon}(f_{\varphi}) - \frac{1}{n}\sum_{l=1}^{n}\hat{\ell}_{f_{\varphi}}(\mathbf{x}^{l}, \mathbf{y}^{l})$ as the following form
\begin{equation}
\begin{split}
 \Big(R_{\varepsilon}(f_{\varphi}) -  \mathbb{E}_{(\mathbf{x}, \mathbf{y}) \sim \hat{\varepsilon}}&{\hat{\ell}}_{f_{\varphi}}(\mathbf{x}, \mathbf{y})\Big)\\
+ \Big(\mathbb{E}_{(\mathbf{x}, \mathbf{y}) \sim \hat{\varepsilon}}{\hat{\ell}}_{f_{\varphi}}(\mathbf{x}, \mathbf{y}) - & \frac{1}{n}\sum_{l=1}^{n}\hat{\ell}_{f_{\varphi}}(\mathbf{x}^{l}, \mathbf{y}^{l})\Big),
\end{split}
\end{equation}
and bound the two terms with Theorem~\ref{suppl_thm:main_1} and Theorem~\ref{suppl_thm:main_2} respectively. We first give Lemma~\ref{suppl_lemma:gaussian_bernoulli} to prove Theorem~\ref{suppl_thm:main_1}.

\begin{lemma}
A random variable $\sigma$ obeys a Bernoulli distribution, where $\mathcal{P}\{\sigma=+1\}=p, \mathcal{P}\{\sigma=-1\}=q\ (p+q =1).$
Another independent random variable $\gamma$ obeys a standard Gaussian distribution where $\gamma \sim N(0,1)$. Then the multiplication $\xi = \sigma\gamma$ still obeys standard Gaussian distribution.
\label{suppl_lemma:gaussian_bernoulli}
\end{lemma}
\noindent\emph{Proof.} $\forall z \in \mathbb{R},$
\begin{equation*}
\begin{split}
&\mathcal{P}\{\xi \leq z\} = \mathcal{P}\{\sigma\gamma \leq z\}\\
= & \mathcal{P}\{\sigma > 0, \gamma \leq \frac{z}{\sigma}\} + \mathcal{P}\{\sigma < 0, \gamma \geq \frac{z}{\sigma}\}\\
= & \mathcal{P}\{\sigma = +1, \gamma \leq z\} + \mathcal{P}\{\sigma = -1, \gamma \geq -z\}\\
\overset{\rm{(\romannumeral 1)}}= & \mathcal{P}\{\sigma = +1\}\mathcal{P}\{\gamma \leq z\} + \mathcal{P}\{\sigma = -1\} \mathcal{P}\{\gamma \geq -z\}\\
= & p\int_{-\infty}^{z}\frac{1}{\sqrt{2\pi}}e^{-\frac{x^2}{2}}dx + q\int_{-z}^{\infty}\frac{1}{\sqrt{2\pi}}e^{-\frac{x^2}{2}}dx.\\
\end{split}
\end{equation*}
(\romannumeral 1) holds due to the independence of $\sigma$ and $\gamma$. The density function of $\xi$ is
\begin{equation*}
\begin{split}
f_{\xi}(z)= &\frac{\mathrm{d} \mathcal{P}\{\xi \leq z\}}{\mathrm{d}z}\\
= & p\frac{\mathrm{d}}{\mathrm{d}z}\int_{-\infty}^{z}\frac{1}{\sqrt{2\pi}}e^{-\frac{x^2}{2}}dx + q\frac{\mathrm{d}}{\mathrm{d}z}\int_{-z}^{\infty}\frac{1}{\sqrt{2\pi}}e^{-\frac{x^2}{2}}dx\\
= & p\frac{1}{\sqrt{2\pi}}e^{-\frac{z^2}{2}} + q(-1)^{2}\frac{1}{\sqrt{2\pi}}e^{-\frac{z^2}{2}}\\
= & \frac{1}{\sqrt{2\pi}}e^{-\frac{z^2}{2}}.\\
\end{split}
\end{equation*}
Therefore, $\xi$ obeys standard Gaussian distribution.$\hfill\blacksquare$

\begin{thm}[\textbf{Theorem~\ref{thm:main_1} in main paper}]
Let $\mathcal{F}$ and $\Pi_1\mathcal{F}(\mathbf{x})$ be the same as in previous theorems. For $\rho >0$, we have
\begin{equation*}
\begin{split}
R_{\varepsilon}(f_{\varphi}) \leq & \mathbb{E}_{(\mathbf{x}, \mathbf{y}) \sim \hat{\varepsilon}}{\hat{\ell}}_{f_{\varphi}}(\mathbf{x}, \mathbf{y})\\
 + & \frac{k\sqrt{2\pi}}{\rho}\mathbb{E}_{\mu \sim \varepsilon}\mathbb{E}_{(\mathbf{x}, \mathbf{y}) \sim \mu^{m}}\Gamma(\Pi_1\mathcal{F}(\mathbf{x})).
\end{split}
\end{equation*}
\label{suppl_thm:main_1}
\end{thm}
\noindent\emph{Proof.} Define vector spaces $\mathcal{F}_{\rho}(\mathbf{x}, \mathbf{y})=\{\big(\rho_{f_{\varphi(\mathbf{x}, \mathbf{y})}}(x_{1},y_{1}), ..., \rho_{f_{\varphi(\mathbf{x}, \mathbf{y})}}(x_{m},y_{m}) \big): \varphi \in \mathcal{D}\}, \Phi_{\rho}{\scalebox{.8}{$\circ$}} \mathcal{F}_{\rho}(\mathbf{x}, \mathbf{y}) = \{\big(\Phi_{\rho}{\scalebox{.8}{$\circ$}}\rho_{f_{\varphi(\mathbf{x}, \mathbf{y})}}(x_{1},y_{1}), ..., \Phi_{\rho}{\scalebox{.8}{$\circ$}}\rho_{f_{\varphi(\mathbf{x}, \mathbf{y})}}(x_{m},y_{m}) \big): \varphi \in \mathcal{D}\}$, we then have
\begin{equation*}
\begin{split}
&R_{\varepsilon}(f_{\varphi})- \mathbb{E}_{(\mathbf{x}, \mathbf{y}) \sim \hat{\varepsilon}}\hat{\ell}_{f_{\varphi}}(\mathbf{x}, \mathbf{y})\\
= & \mathbb{E}_{\mu \sim \varepsilon}\mathbb{E}_{(\mathbf{x}, \mathbf{y}) \sim \mu^{m}}\big[ \mathbb{E}_{(x,y)\sim \mu}\ell_{f_{\varphi(\mathbf{x}, \mathbf{y})}}(x,y)-\hat{\ell}_{f_{\varphi}}(\mathbf{x}, \mathbf{y})\big]\\
\leq & \mathbb{E}_{\mu \sim \varepsilon}\mathbb{E}_{(\mathbf{x}, \mathbf{y}) \sim \mu^{m}}\big[ \sup_{\varphi \in \mathcal{D}}\bigg(\mathbb{E}_{(x,y)\sim \mu}\ell_{f_{\varphi(\mathbf{x}, \mathbf{y})}}(x,y)-\hat{\ell}_{f_{\varphi}}(\mathbf{x}, \mathbf{y})\bigg)\big]\\
= & \mathbb{E}_{\mu \sim \varepsilon}\mathbb{E}_{(\mathbf{x}, \mathbf{y}) \sim \mu^{m}}\big[ \sup_{\varphi \in \mathcal{D}}\bigg( \mathbb{E}_{(x,y)\sim \mu}\Phi_{\rho}{\scalebox{.8}{$\circ$}}\rho_{f_{\varphi(\mathbf{x}, \mathbf{y})}}(x,y)\\
& -\frac{1}{m}\sum_{i=1}^{m}\Phi_{\rho}{\scalebox{.8}{$\circ$}}\rho_{f_{\varphi(\mathbf{x}, \mathbf{y})}}(x_{i}, y_{i})\bigg)\big]\\
 \overset{\rm{(\romannumeral 1)}}  \leq & \sqrt{\frac{\pi}{2}}\mathbb{E}_{\mu \sim \varepsilon}\mathbb{E}_{(\mathbf{x}, \mathbf{y}) \sim \mu^{m}}\Gamma(\Phi {\scalebox{.8}{$\circ$}} \mathcal{F}_{\rho}(\mathbf{x}, \mathbf{y}))\\
\overset{\rm{(\romannumeral 2)}} \leq & \sqrt{\frac{\pi}{2}}\mathbb{E}_{\mu \sim \varepsilon}\mathbb{E}_{(\mathbf{x}, \mathbf{y}) \sim \mu^{m}}\frac{1}{\rho}\Gamma(\mathcal{F}_{\rho}(\mathbf{x}, \mathbf{y})).\\
\end{split}
\end{equation*}
The inequality (\romannumeral 1) holds due to Theorem~\ref{suppl_thm:wellner_bartlett}, (\romannumeral 1) and (\romannumeral 3). And the inequality (\romannumeral 2) uses Eq.~(\ref{suppl_eq:talagrand}). Using the sub-additivity of the $\sup$ function , we can bound the Gaussian complexity
\begin{equation}
\begin{split}
&\Gamma(\mathcal{F}_{\rho}(\mathbf{x}, \mathbf{y}))= \mathbb{E}_{\gamma}\sup_{\varphi \in \mathcal{D}}\frac{2}{m}\sum_{i=1}^{m}\rho_{f_{\varphi(\mathbf{x}, \mathbf{y})}}(x_i, y_i)\gamma_i\\
= & \mathbb{E}_{\gamma}\sup_{\varphi \in \mathcal{D}}\frac{2}{m}\sum_{i=1}^{m}\big( f_{\varphi(\mathbf{x}, \mathbf{y})}(x_i,y_i) - \max_{y \neq y_i}f_{\varphi(\mathbf{x}, \mathbf{y})}(x_i,y)\big)\gamma_i\\
\leq & \mathbb{E}_{\gamma}\sup_{\varphi \in \mathcal{D}}\frac{2}{m}\sum_{i=1}^{m}f_{\varphi(\mathbf{x}, \mathbf{y})}(x_i,y_i)\gamma_i \\
&+ \mathbb{E}_{\gamma}\sup_{\varphi \in \mathcal{D}}\frac{2}{m}\sum_{i=1}^{m} - \max_{y \neq y_i}f_{\varphi(\mathbf{x}, \mathbf{y})}(x_i,y)\gamma_i.\\
\end{split}
\label{suppl_eq:main_1_gaussian_comlexity}
\end{equation}
With the sub-additivity of $\sup$, the first term of the above result can be bounded by
\begin{equation}
\begin{split}
&\mathbb{E}_{\gamma}\sup_{\varphi \in \mathcal{D}}\frac{2}{m}\sum_{i=1}^{m}f_{\varphi(\mathbf{x}, \mathbf{y})}(x_i,y_i)\gamma_i\\
=& \mathbb{E}_{\gamma}\sup_{\varphi \in \mathcal{D}}\frac{2}{m}\sum_{i=1}^{m}\sum_{y \in \mathcal{Y}}\gamma_i f_{\varphi(\mathbf{x}, \mathbf{y})}(x_i,y)\mathbbm{1}_{y=y_i}\\
\overset{\rm{(\romannumeral 1)}}\leq & \sum_{y \in \mathcal{Y}}\mathbb{E}_{\gamma}\sup_{\varphi \in \mathcal{D}}\frac{2}{m}\sum_{i=1}^{m}\gamma_i f_{\varphi(\mathbf{x}, \mathbf{y})}(x_i,y)\big(\frac{\epsilon_i+1}{2}\big)\\
\leq & \sum_{y \in \mathcal{Y}}\mathbb{E}_{\gamma}\sup_{\varphi \in \mathcal{D}}\frac{2}{m}\sum_{i=1}^{m}f_{\varphi(\mathbf{x}, \mathbf{y})}(x_i,y)\frac{\epsilon_i\gamma_i}{2} \\
& + \sum_{y \in \mathcal{Y}}\mathbb{E}_{\gamma}\sup_{\varphi \in \mathcal{D}}\frac{2}{m}\sum_{i=1}^{m}f_{\varphi(\mathbf{x}, \mathbf{y})}(x_i,y)\frac{\gamma_i}{2}\\
= & k\Gamma(\Pi_{1}\mathcal{F}(\mathbf{x})).
\end{split}
\label{suppl_eq:main_1_gaussian_comlexity_1}
\end{equation}
Inequality (\romannumeral 1) uses the fact that $\epsilon_i = 2\mathbbm{1}_{y=y_i}-1 \in \{-1,+1\}$. The last inequality holds because from Lemma~\ref{suppl_lemma:gaussian_bernoulli} we know $\epsilon_i\gamma_i$ and $\gamma_i$ admits the same distribution, and $|\mathcal{Y}|=k$. Similarly, we can obtain the upper bound of the second term in the r.h.s of Eq.~(\ref{suppl_eq:main_1_gaussian_comlexity})
\begin{equation}
\begin{split}
&\mathbb{E}_{\gamma}\sup_{\varphi \in \mathcal{D}}\frac{2}{m}\sum_{i=1}^{m} - \max_{y \neq y_i}f_{\varphi(\mathbf{x}, \mathbf{y})}(x_i,y)\gamma_i\\
\leq &\mathbb{E}_{\gamma}\sup_{\varphi \in \mathcal{D}}\frac{2}{m}\sum_{i=1}^{m} \max_{y \in \mathcal{Y}}f_{\varphi(\mathbf{x}, \mathbf{y})}(x_i,y)\gamma_i\\
\leq &\sum_{y \in \mathcal{Y}}\mathbb{E}_{\gamma}\sup_{\varphi \in \mathcal{D}}\frac{2}{m}\sum_{i=1}^{m}f_{\varphi(\mathbf{x}, \mathbf{y})}(x_i,y)\gamma_i \qquad \rm{(Theorem~\ref{suppl_thm:max_f})}\\
= & k\Gamma(\Pi_{1}\mathcal{F}(\mathbf{x})).
\end{split}
\label{suppl_eq:main_1_gaussian_comlexity_2}
\end{equation}
Combining Eqs.~(\ref{suppl_eq:main_1_gaussian_comlexity})-(\ref{suppl_eq:main_1_gaussian_comlexity_2}), we derive the expected result. $\hfill\blacksquare$

\begin{thm}[\textbf{Theorem~\ref{thm:main_2} in main paper}]
Let $\mathcal{F}$ and $\Pi_1\mathcal{F}(\mathbf{X})$ be the same as in previous theorems. For any $\delta > 0$, we have, with probability at least $1-\delta$ on the draw of the sample $((\mathbf{x}^{1},\mathbf{y}^{1}),...,(\mathbf{x}^{n},\mathbf{y}^{n}))$,
\begin{equation*}
\begin{split}
\mathbb{E}_{(\mathbf{x}, \mathbf{y}) \sim \hat{\varepsilon}}{\hat{\ell}}_{f_{\varphi}}(\mathbf{x}, \mathbf{y}) \leq & \frac{1}{n}\sum_{l=1}^{n}\hat{\ell}_{f_{\varphi}}(\mathbf{x}^{l}, \mathbf{y}^{l})  + \sqrt{\frac{\ln(1/\delta)}{2n}}\\
+ & \frac{k\sqrt{2m\pi}}{\rho}\mathbb{E}_{(\mathbf{X},\mathbf{Y}) \sim \hat{\varepsilon}^{n}}\Gamma(\Pi_{1}\mathcal{F}(\mathbf{X})).
\end{split}
\end{equation*}
\label{suppl_thm:main_2}
\end{thm}
\noindent\emph{Proof.} Fix a meta-sample $(\mathbf{X},\mathbf{Y})=((\mathbf{x^{1}},\mathbf{y}^{1}),...,(\mathbf{x}^{n},\mathbf{y}^{n}))$. Define Gaussian processes $\Omega_{\varphi}$ and $\Xi_{\varphi}$ indexed by ${\varphi}$ as follows:
\begin{equation*}
\Omega_{\varphi}\hspace{-0.05in}=\hspace{-0.05in}\sum_{l=1}^{n}\gamma^{l}\hat{\ell}_{f_{\varphi}}(\mathbf{x}^{l},\mathbf{y}^{l}) \quad \Xi_{\varphi}\hspace{-0.05in}=\hspace{-0.05in}\sum_{l=1}^{n}\sum_{i=1}^{m}\frac{\gamma_i^{l}}{\sqrt{m}\rho}\rho_{f_{\varphi(\mathbf{x}^{l}, \mathbf{y}^{l})}}(x_i^{l},y_i^{l}),
\end{equation*}
where $\gamma^{l}$ and $\gamma_{i}^{l}$ are mutually independent standard Gaussian distributed variables. Define function class $\mathcal{G}_{\varphi}=\{(\mathbf{x}, \mathbf{y}) \mapsto \hat{\ell}_{f_{\varphi}}(\mathbf{x}, \mathbf{y})\}$ and observe that $(2/n)\mathbb{E}\sup_{\varphi\in \mathcal{D}}\Omega_{\varphi}=\Gamma(\mathcal{G}_{\varphi}(\mathbf{X},\mathbf{Y}))$. Then $\forall \varphi_1, \varphi_2 \in \mathcal{D}$ by using the orthogonality of the $\gamma^{l}$ we have
\begin{small}
\begin{equation*}
\begin{split}
& \mathbb{E}(\Omega_{\varphi_1}-\Omega_{\varphi_2})^{2}\\
= &\mathbb{E}_{\gamma}\bigg(\sum_{l=1}^{n}\gamma^{l}\big({\hat{\ell}}_{\rho_{f_{\varphi_1}}}(\mathbf{x}^{l},\mathbf{y}^{l})-{\hat{\ell}}_{\rho_{f_{\varphi_2}}}(\mathbf{x}^{l},\mathbf{y}^{l})\big)\bigg)^{2}\\
= & \sum_{l=1}^{n}\big({\hat{\ell}}_{\rho_{f_{\varphi_1}}}(\mathbf{x}^{l},\mathbf{y}^{l})-{\hat{\ell}}_{\rho_{f_{\varphi_2}}}(\mathbf{x}^{l},\mathbf{y}^{l})\big)^{2}\\
= & \sum_{l=1}^{n} \bigg(\frac{1}{m}\sum_{i=1}^{m}\Phi_{\rho}(\rho_{f_{\varphi_1(\mathbf{x}^{l},\mathbf{y}^{l})}}(x_i^{l}, y_i^{l}))-\Phi_{\rho}(\rho_{f_{\varphi_2(\mathbf{x}^{l},\mathbf{y}^{l})}}(x_i^{l}, y_i^{l}))\bigg)^{2}\\
\leq & \sum_{l=1}^{n} \bigg(\frac{1}{m}\sum_{i=1}^{m}|\Phi_{\rho}(\rho_{f_{\varphi_1(\mathbf{x}^{l},\mathbf{y}^{l})}}(x_i^{l}, y_i^{l}))-\Phi_{\rho}(\rho_{f_{\varphi_2(\mathbf{x}^{l},\mathbf{y}^{l})}}(x_i^{l}, y_i^{l}))|\bigg)^{2}\\
\overset{\rm{(\romannumeral 1)}}\leq & \sum_{l=1}^{n} \frac{1}{m^{2}\rho^{2}}\bigg(\sum_{i=1}^{m}|\rho_{f_{\varphi_1(\mathbf{x}^{l},\mathbf{y}^{l})}}(x_i^{l}, y_i^{l})-\rho_{f_{\varphi_2(\mathbf{x}^{l},\mathbf{y}^{l})}}(x_i^{l}, y_i^{l})|\bigg)^{2}\\
\overset{\rm{(\romannumeral 2)}}\leq & \sum_{l=1}^{n} \frac{1}{m\rho^{2}}\sum_{i=1}^{m}\big(\rho_{f_{\varphi_1(\mathbf{x}^{l},\mathbf{y}^{l})}}(x_i^{l}, y_i^{l})-\rho_{f_{\varphi_2(\mathbf{x}^{l},\mathbf{y}^{l})}}(x_i^{l}, y_i^{l})\big)^{2}\\
= & \mathbb{E}(\Xi_{\varphi_1}-\Xi_{\varphi_2})^{2}
\end{split}
\end{equation*}
\end{small}
Inequality (\romannumeral 1) uses the Lipschitz Property of margin loss and inequality applies Mean Value Inequality. Then from Theorem~\ref{suppl_thm:slepian} we have $\mathbb{E}_{\varphi \in \mathcal{D}}\Omega_{\varphi} \leq \mathbb{E}_{\varphi \in \mathcal{D}}\Xi_{\varphi}$. Multiplying with $2/n$ this becomes
\begin{equation*}
\begin{split}
\Gamma(\mathcal{G}_{\varphi}(\mathbf{X},\mathbf{Y})) & \leq \frac{2}{n}\mathbb{E}_{f \in \mathcal{F}}\sum_{l=1}^{n}\sum_{i=1}^{m}\frac{\gamma_i^{l}}{\sqrt{m}\rho}\rho_{f_{\varphi(\mathbf{x}^{l}, \mathbf{y}^{l})}}(x_i^{l},y_i^{l})\\
&= \frac{\sqrt{m}}{\rho}\Gamma(\mathcal{F}_{\rho}(\mathbf{X}, \mathbf{Y}))
\end{split}
\end{equation*}
where $\mathcal{F}_{\rho}(\mathbf{X}, \mathbf{Y})= \{\big(\rho_{f_{\varphi_{(\mathbf{X},\mathbf{Y})}}}(x_{1}^{1},y_{1}^{1}),...,\rho_{f_{\varphi_{(\mathbf{X},\mathbf{Y})}}}(x_{m}^{1},y_{m}^{1})\\
,...,\rho_{f_{\varphi_{(\mathbf{X},\mathbf{Y})}}}(x_{1}^{n},y_{1}^{n}),...,
\rho_{f_{\varphi_{(\mathbf{X},\mathbf{Y})}}}(x_{m}^{n},y_{m}^{n})\big): \varphi \in \mathcal{D}\}$ and we define the scoring function $\rho_{f_{\varphi_{(\mathbf{X},\mathbf{Y})}}}(x_{i}^{l},y_{i}^{l})= \rho_{f_{\varphi(\mathbf{x}^{l},\mathbf{y}^{l})}}(x_{i}^{l},y_{i}^{l}), \forall i \in [m], l \in [n]$.
Analogous to the demonstration process to bound the Gaussian complexity $\Gamma(\mathcal{F}_{\rho}(\mathbf{x}, \mathbf{y}))$ in Theorem~\ref{suppl_thm:main_1}, we can bound $\Gamma(\mathcal{F}_{\rho}(\mathbf{X}, \mathbf{Y}))$ with $k\Gamma(\Pi_{1}\mathcal{F}(\mathbf{X}))$ and draw the conclusion
\begin{equation}
\Gamma(\mathcal{G}_{\varphi}(\mathbf{X},\mathbf{Y})) \leq \frac{2k\sqrt{m}}{\rho}\Gamma(\Pi_{1}\mathcal{F}(\mathbf{X})).
\label{suppl_eq:main_2_gaussian_complexity}
\end{equation}
Combining Eq.~(\ref{suppl_eq:main_2_gaussian_complexity}) and Theorem~\ref{suppl_thm:wellner_bartlett} (\romannumeral 2), (\romannumeral 3) completes the proof. $\hfill\blacksquare$

\section{Covering Number Transfer Bound}
\label{suppl_sect:covering_number_bound}

\begin{thm}[Refined Dudley Entropy Bound, Theorem~\ref{thm:dudley_entropy_bound} in main paper]
For any real-valued function class $\mathcal{F}$ containing function $f: \mathcal{X} \rightarrow \mathbb{R}$, assume that $\sup_{f \in \mathcal{F}}\|f\|_{2}$ is bounded under the $\mathcal{L}_{2}(\mathbf{x})$ and $\mathcal{L}_{2}(\mathbf{X})$ metric respectively. Then
\begin{equation*}
\begin{split}
\Gamma(\mathcal{F}(\mathbf{x})) \leq \frac{24}{\sqrt{m}}\int_{0}^{\sup_{f \in \mathcal{F}}\|f\|_{2}}\sqrt{\log{\mathcal{N}(\tau, \mathcal{F}, \mathcal{L}_{2}(\mathbf{x}))}}\mathrm{d}\tau,\\
\Gamma(\mathcal{F}(\mathbf{X})) \leq \frac{24}{\sqrt{nm}}\int_{0}^{\sup_{f \in \mathcal{F}}\|f\|_{2}}\sqrt{\log{\mathcal{N}(\tau, \mathcal{F}, \mathcal{L}_{2}(\mathbf{X}))}}\mathrm{d}\tau.
\end{split}
\end{equation*}
\label{suppl_thm:dudley_entropy_bound}
\end{thm}
\noindent\emph{Proof.} We just prove the first part. The main idea of the proof is to use generic chaining technique. Let $\alpha_{0}=\sup_{f \in \mathcal{F}}\|f\|_{2}, \alpha_{i}=2^{-i}\sup_{f \in \mathcal{F}}\|f\|_{2} (i \geq 1)$. $T_{0}=\{0\}$ is the $\alpha_{0}$-cover of $\mathcal{F}$, and let $T_{i} (i \geq 1)$ be an $\alpha_{i}$-cover of $\mathcal{F}$ with the smallest cardinality. Then $\forall i \geq 0$, we choose $\hat{f}_{i}$ from $T_{i}$, such that we have $\|f - \hat{f}_{i}\|_{2} \leq \alpha_{i}$ and can rewrite $f$ as a `chain' $f = f- \hat{f}_{N} + \sum_{i=1}^{N}(\hat{f}_{i}-\hat{f}_{i-1})$.
\begin{equation}
\begin{split}
& \Gamma(\mathcal{F}(\mathbf{x})) = \frac{2}{m}\mathbb{E}_{\bm{\gamma}}\sup_{f \in \mathcal{F}}\sum_{i=1}^{m}\gamma_{i}f(x_i) = \frac{2}{m}\mathbb{E}_{\bm{\gamma}}\sup_{f \in \mathcal{F}} \langle\bm{\gamma}, f(\mathbf{x})\rangle\\
= & \frac{2}{m}\mathbb{E}_{\bm{\gamma}}\sup_{f \in \mathcal{F}} \Big(\langle\bm{\gamma}, f - \hat{f}_{N}\rangle + \langle\bm{\gamma}, \sum_{i=1}^{N}\hat{f}_{i}-\hat{f}_{i-1}\rangle \Big)\\
 \leq & \frac{2}{m}\mathbb{E}_{\bm{\gamma}}\sup_{f \in \mathcal{F}} \langle\bm{\gamma}, f - \hat{f}_{N}\rangle + \sum_{i=1}^{N}\frac{2}{m}\mathbb{E}_{\bm{\gamma}}\sup_{f \in \mathcal{F}} \langle\bm{\gamma}, \hat{f}_{i}-\hat{f}_{i-1}\rangle. \\
\end{split}
\label{suppl_eq:refined_dudley_bound}
\end{equation}
To bound the first term of the above equation, we have
\begin{equation}
\begin{split}
&\frac{2}{m}\mathbb{E}_{\bm{\gamma}}\sup_{f \in \mathcal{F}} \langle\bm{\gamma}, f - \hat{f}_{N}\rangle =  \frac{2}{m}\mathbb{E}_{\bm{\gamma}}\sup_{f \in \mathcal{F}}\sum_{i=1}^{m}\gamma_{i}\big (f(x_i)-\hat{f}_{N}(x_i) \big)\\
\overset{\rm{(\romannumeral 1)}}\leq & \frac{2}{m}\mathbb{E}_{\bm{\gamma}}\sup_{f \in \mathcal{F}}\big(\sum_{i=1}^{m}\gamma_{i}^{2}\big)^{\frac{1}{2}}\big (\sum_{i=1}^{m} \big (f(x_i)-\hat{f}_{N}(x_i)\big)^{2} \big)^{\frac{1}{2}}\\
= & 2\big(\mathbb{E}_{\bm{\gamma}}\|\bm{\gamma}\|_{2}\big) \big(\sup_{f \in \mathcal{F}}\big\|f - \hat{f}_{N}\|_{2}\big)\\
\overset{\rm{(\romannumeral 2)}}\leq & 2\big(\sqrt{\mathbb{E}_{\bm{\gamma}}\|\bm{\gamma}\|_{2}^{2}}\big) \big( \sup_{f \in \mathcal{F}}\big\|f - \hat{f}_{N}\|_{2} \big)\leq  2\alpha_{N}.
\label{suppl_eq:refined_dudley_bound_1}
\end{split}
\end{equation}
(\romannumeral 1) and (\romannumeral 2) use Cauchy-Schwarz and Jensen inequality respectively. The last inequality of Eq.~(\ref{suppl_eq:refined_dudley_bound_1}) holds because $\mathbb{E}_{\bm{\gamma}}\|\bm{\gamma}\|_{2}^{2}=\mathbb{E}\frac{1}{m}\sum_{i=1}^{m}\gamma_{i}^2=1$ and $T_{N}$ is a $\alpha_{N}$-cover of $\mathcal{F}$. To bound the second term in the r.h.s. of Eq.~(\ref{suppl_eq:refined_dudley_bound}), with triangle inequality we have $\|\hat{f}_{i}-\hat{f}_{i-1}\|_{2} \leq \|\hat{f}_{i}-f\|_{2} + \|f - \hat{f}_{i-1}\|_{2} \leq \alpha_{i}+\alpha_{i-1} = 3\alpha_{i}$. Define function class $\hat{\mathcal{F}}_{i}=\{\hat{f}_{i}-\hat{f}_{i-1}: \hat{f}_{i} \in T_{i}, \hat{f}_{i-1} \in T_{i-1}\}$, then we have
\begin{equation}
\begin{split}
&\sum_{i=1}^{N}\frac{2}{m}\mathbb{E}_{\bm{\gamma}}\sup_{f \in \mathcal{F}} \langle\bm{\gamma}, \hat{f}_{i}-\hat{f}_{i-1}\rangle = \sum_{i=1}^{N}\Gamma(\hat{\mathcal{F}}_{i})\\
\leq & \sum_{i=1}^{N}3 \alpha_{i}\frac{2\sqrt{2\log{|T_{i}|\cdot|T_{i-1}|}}}{\sqrt{m}} \qquad \rm{(Lemma\ \ref{suppl_lemma:massart})}\\
\leq & \sum_{i=1}^{N}12 \alpha_{i}\frac{\sqrt{\log{|T_{i}|}}}{\sqrt{m}} = \frac{24}{\sqrt{m}}\sum_{i=1}^{N}\big( \alpha_{i} - \alpha_{i+1}\big)\sqrt{\log{|T_{i}|}}\\
\leq & \frac{24}{\sqrt{m}}\sum_{i=1}^{N}\big( \alpha_{i} - \alpha_{i+1}\big)\sqrt{\log{\mathcal{N}(\tau, \mathcal{F}, \mathcal{L}_{2}(\mathbf{x}))}} \qquad \rm{(\tau < \alpha_{i})}\\
\leq & \frac{24}{\sqrt{m}}\sum_{i=1}^{N}\int_{\alpha_{i+1}}^{\alpha_{i}}\sqrt{\log{\mathcal{N}(\tau, \mathcal{F}, \mathcal{L}_{2}(\mathbf{x}))}}\mathrm{d}\tau\\
= & \frac{24}{\sqrt{m}}\int_{\alpha_{N+1}}^{\alpha_{1}}\sqrt{\log{\mathcal{N}(\tau, \mathcal{F}, \mathcal{L}_{2}(\mathbf{x}))}}\mathrm{d}\tau.\\
\end{split}
\label{suppl_eq:refined_dudley_bound_2}
\end{equation}
$\forall \epsilon > 0$, let $N = \sup\{i: \alpha_{i} > 2\epsilon\}$. Then we have $\epsilon < \alpha_{N+1} \leq 2\epsilon$ and $\alpha_{N} = 2 \alpha_{N+1} \leq 4\epsilon$. Combining Eqs.~(\ref{suppl_eq:refined_dudley_bound})-(\ref{suppl_eq:refined_dudley_bound_2}) and recall that $\alpha_{0}=\sup_{f \in \mathcal{F}}\|f\|_{2}$, we have
\begin{equation}
\Gamma(\mathcal{F}(\mathbf{x})) \leq 8\epsilon + \frac{24}{\sqrt{m}}\int_{\epsilon}^{\sup_{f \in \mathcal{F}}\|f\|_{2}}\sqrt{\log{\mathcal{N}(\tau, \mathcal{F}, \mathcal{L}_{2}(\mathbf{x}))}}\mathrm{d}\tau.
\label{suppl_eq:refined_dudley_bound_3}
\end{equation}
Let $\epsilon \rightarrow 0$ in the right hand side of Eq.~(\ref{suppl_eq:refined_dudley_bound_3}), we obtain the final result. $\hfill\blacksquare$\\
Combining Theorem~\ref{suppl_thm:main} and Theorem~\ref{suppl_thm:main_covering_number} we immediately obtain the following margin-based covering number bound for few-shot learning.
\begin{thm}[\textbf{Margin-based Transfer Bound for MLMC with Covering Number, Theorem~\ref{thm:main_covering_number} in main paper}]
Let $\mathcal{F}$ and $\Pi_1\mathcal{F}$ be the same as in previous theorems. Given a classification algorithm $f$ and a margin parameter $\rho >0$, for any environment $\varepsilon \in M_1(M_1(\mathcal{X} \times \mathcal{Y}))$ and for any $\delta > 0$, with probability at least $1-\delta$ on the data $(\mathbf{X}, \mathbf{Y}) \sim {\hat{\varepsilon}}^{n}$, we have for all feature maps $\varphi \in \mathcal{D}$ that
\begin{equation*}
\begin{split}
& R_{\varepsilon}(f_{\varphi})\leq  \frac{1}{n}\sum_{l=1}^{n}\hat{\ell}_{f_{\varphi}}(\mathbf{x}^{l}, \mathbf{y}^{l})+ \sqrt{\frac{\ln(1/\delta)}{2n}}\\
& +   \frac{24k\sqrt{2\pi}}{\rho\sqrt{n}}\mathbb{E}_{(\mathbf{X},\mathbf{Y}) \sim \hat{\varepsilon}^{n}}\int_{0}^{L}\sqrt{\log{\mathcal{N}(\tau, \Pi_{1}\mathcal{F}, \mathcal{L}_{2}(\mathbf{X}))}}\mathrm{d}\tau\\
& +  \hspace{-0.02in} \frac{24k\sqrt{2\pi}}{\rho\sqrt{m}}\mathbb{E}_{\mu \sim \varepsilon}\mathbb{E}_{(\mathbf{x}, \mathbf{y}) \sim \mu^{m}} \hspace{-0.03in} \int_{0}^{L}\hspace{-0.05in} \sqrt{\log{\mathcal{N}(\tau, \Pi_{1}\mathcal{F}, \mathcal{L}_{2}(\mathbf{x}))}}\mathrm{d}\tau.\\
\end{split}
\end{equation*}
\label{suppl_thm:main_covering_number}
\end{thm}

\section{VC-dimension Transfer Bound}
\label{suppl_sect:transfer_bound_fsl}

In this section, we give the bound of the covering number $\mathcal{N}(\Pi_{1}\mathcal{F}(\mathbf{X}))$ and $\mathcal{N}(\Pi_{1}\mathcal{F}(\mathbf{x}))$ in Theorem~\ref{suppl_thm:main_covering_number} with the VC-dimension of the hypothesis space $\Pi_{1}\mathcal{F}$. Then we yield our most import theoretical result in Theorem~\ref{suppl_thm:transfer_bound_fsl}.
\begin{thm}[Theorem 2.6.7 in \cite{vaart1996springer}]
Let $\mathcal{F}$ be a real-valued function class on $\mathcal{X}$ with VC-dimension $v$. Assume that $\mathcal{F}$ is uniformly bounded by $b>0$. Then, for any probability distribution $Q$ on $\mathcal{X}$ and for $p \geq 1$
\begin{equation*}
\mathcal{N}(\tau, \mathcal{F}, \|\cdot\|_{L_{p}(Q)}) \leq C_{0}(v+1)(16e)^{v+1}(\frac{b}{\tau})^{pv},
\end{equation*}
\label{suppl_thm:covering_number_vc_bound}
\end{thm}
for some uniform constant $C_{0} >0$, where for any $f, g \in \mathcal{F}$, $\|f-g\|_{L_{p}(Q)}=(\int |f-g|^{p}dQ)^{1/p}$.

\begin{table*}[t]
\caption{The 5-way $s$-shot classification results on the \textbf{miniImageNet $\rightarrow$ CUB} dataset. We report the average accuracy (\%, top-1) as well as the 95\% confidence interval over all 600 test episodes. We compare the \textbf{Multi-Margin} loss with the \textbf{Cross-Entropy} loss. }
\scalebox{0.78}{
\begin{tabular}{l|cc|cc|cc}
\toprule[2pt]
     & \multicolumn{2}{|c}{\textbf{5-way 5-shot}} & \multicolumn{2}{|c}{\textbf{5-way 10-shot}} & \multicolumn{2}{|c}{\textbf{5-way 20-shot}}\\
    \textbf{Model} &\textbf{Cross-Entropy} & \textbf{Multi-Margin} & \textbf{Cross-Entropy} & \textbf{Multi-Margin} & \textbf{Cross-Entropy} & \textbf{Multi-Margin}\\
    \midrule[1pt]
    Baseline++ \cite{yu2019iclr} & $66.21\pm0.70$ & $67.90\pm0.69$ & $75.45\pm0.61$ & $76.48\pm0.60$ & $81.18\pm0.53$ & $83.04\pm0.54$\\
    MAML \cite{finn2017icml} & $61.09\pm0.57$ & $60.85\pm0.54$ & $70.54\pm0.74$ & $69.67\pm0.77$ & $76.65\pm0.72$ & $75.87\pm0.75$\\
    \midrule[1pt]
    MatchingNet \cite{oriol2016nips} & $64.47\pm0.73$ & $63.34\pm0.71$ & $72.03\pm0.71$ & $71.72\pm0.69$ & $78.60\pm0.61$ & $77.48\pm0.61$\\
    ProtoNet \cite{jake2017nips} & $64.51\pm0.75$  & $64.41\pm0.75$ & $72.62\pm0.66$ & $72.61\pm0.71$ & $79.63\pm0.63$ & $79.79\pm0.60$\\
    RelationNet \cite{sung2018cvpr} & $64.68\pm0.74$ & $65.16\pm0.74$ & $69.82\pm0.67$ & $70.83\pm0.71$ & $75.62\pm0.65$ & $75.82\pm0.61$ \\
    MetaOptNet \cite{lee2019cvpr} & $65.17\pm0.73$ & $64.62\pm0.74$ & $74.64\pm0.68$ & $73.89\pm0.67$ & $80.45\pm0.63$ & $80.24\pm0.65$\\
    \bottomrule[2pt]
  \end{tabular}
}
\label{tab:suppl_main_mini}
\end{table*}

\begin{thm}[\textbf{Margin-based Transfer Bound for MLMC with VC-dimension, Theorem~\ref{thm:transfer_bound_fsl} in main paper}]
Let the VC-dimension of $\Pi_1\mathcal{F}$ defined in previous theorems be $v$, and $\Pi_1\mathcal{F}$ is uniformly bounded by $b>0$. Given a classification algorithm $f$ and a margin parameter $\rho >0$, for any environment $\varepsilon \in M_1(M_1(\mathcal{X} \times \mathcal{Y}))$ and for any $\delta > 0$, with probability at least $1-\delta$ on the data $(\mathbf{X}, \mathbf{Y}) \sim {\hat{\varepsilon}}^{n}$, we have for all feature maps $\varphi \in \mathcal{D}$ that
\begin{equation*}
\begin{split}
R_{\varepsilon}(f_{\varphi})\leq & \frac{1}{n}\sum_{l=1}^{n}\hat{\ell}_{f_{\varphi}}(\mathbf{x}^{l}, \mathbf{y}^{l}) + \sqrt{\frac{\ln(1/\delta)}{2n}}\\
+  & (\frac{k}{\rho\sqrt{m}}+\frac{k}{\rho\sqrt{n}})(C_1 \sqrt{v} + C_2),
\end{split}
\end{equation*}
where constants $C_1 = 24\sqrt{2\pi}b(1+\sqrt{\log (16e)}+2\sqrt{2})$ and $C_2 = 24\sqrt{2\pi}b(\sqrt{\log C_0} + \sqrt{\log (16e)})$. $C_0$ is the uniform constant in Theorem~\ref{suppl_thm:covering_number_vc_bound}.
\label{suppl_thm:transfer_bound_fsl}
\end{thm}
\noindent\emph{Proof.}
Notice that for all $y \in \mathcal{Y}$
\begin{equation*}
\sup_{f \in \Pi_1\mathcal{F}}\|f\|_{2} = \sup_{f \in \Pi_1\mathcal{F}} \sqrt{\frac{1}{m}\sum_{i=1}^{m}f(x_i, y)^{2}} \leq \sqrt{\frac{1}{m}\sum_{i=1}^{m}b^2} = b.
\end{equation*}
From Theorem~\ref{suppl_thm:covering_number_vc_bound}, we know there exits a uniform constant $C_0$ such that
\begin{equation*}
\mathcal{N}(\tau, \Pi_{1}\mathcal{F}, \mathcal{L}_{2}(\mathbf{x})) \leq C_0(v+1)(16e)^{v+1}(\frac{b}{\tau})^{2v}.
\end{equation*}
Then the integral of the $\log$ covering number in Theorem~\ref{suppl_thm:main_covering_number} can be bounded by
\begin{equation*}
\begin{split}
& I =\int_{0}^{\sup_{f \in \Pi_1\mathcal{F}}\|f\|_{2}}\sqrt{\log \mathcal{N}(\tau, \Pi_{1}\mathcal{F}, \mathcal{L}_{2}(\mathbf{x}))}\mathrm{d}\tau\\
\leq &\int_{0}^{b}\sqrt{\log{C_0(v+1)(16e)^{v+1}(\frac{b}{\tau})^{2v}}}\mathrm{d}\tau\\
= &\int_{0}^{b}\hspace{-0.05in}\sqrt{\log C_0 + \log (v+1) + (v+1)\log (16e) + 2v\log (\frac{b}{\tau})}\mathrm{d}\tau\\
\overset{\rm{(\romannumeral 1)}}\leq &\int_{0}^{b}\sqrt{\log C_0 + v + (v+1)\log (16e) + 2v\frac{b}{\tau}}\mathrm{d}\tau\\
\overset{\rm{(\romannumeral 2)}} \leq &\int_{0}^{b}\sqrt{\log C_0 + v + (v+1)\log (16e)} + \sqrt{2v\frac{b}{\tau}}\mathrm{d}\tau\\
\leq & \alpha\sqrt{v}+\beta,
\end{split}
\end{equation*}
(\romannumeral 1) and (\romannumeral 2) hold due to the basic inequalities $\ln(x+1)\leq x$ and $\sqrt{x+y}\leq\sqrt{x}+\sqrt{y}$. Further, $\alpha=(1+\sqrt{\log(16e)}+2\sqrt{2})b, \beta= (\sqrt{\log C_0} + \sqrt{\log (16e)})b$. Similarly we can give the same bound for the integral: $\int_{0}^{\sup_{f \in \Pi_1\mathcal{F}}\|f\|_{2}}\sqrt{\mathcal{N}(\tau, \Pi_{1}\mathcal{F}, \mathcal{L}_{2}(\mathbf{X}))}\mathrm{d}\tau \leq \alpha\sqrt{v}+\beta$. Then combining with Theorem~\ref{suppl_thm:main_covering_number}, we have with probability at least $1-\delta$ on the data $(\mathbf{X}, \mathbf{Y}) \sim {\hat{\varepsilon}}^{n}$
\begin{equation*}
\begin{split}
R_{\varepsilon}(f_{\varphi})\leq & \frac{1}{n}\sum_{l=1}^{n}\hat{\ell}_{f_{\varphi}}(\mathbf{x}^{l}, \mathbf{y}^{l}) + \sqrt{\frac{\ln(1/\delta)}{2n}}\\
+  & (\frac{k}{\rho\sqrt{m}}+\frac{k}{\rho\sqrt{n}})(C_1 \sqrt{v} + C_2),
\end{split}
\end{equation*}
where $C_1 = 24\sqrt{2\pi}b(1+\sqrt{\log (16e)}+2\sqrt{2})$ and $C_2 = 24\sqrt{2\pi}b(\sqrt{\log C_0} + \sqrt{\log (16e)})$. $\hfill\blacksquare$\\

\section{Cross-Domain Experiment Results}
\label{suppl_sect:cross-domain mlmc}
We provide more experiment results on the cross-domain dataset \textbf{miniImageNet $\rightarrow$ CUB} in Table~\ref{tab:suppl_main_mini}. Different 5-way classification settings are considered here. We can still find that the results obtained with the multi-margin loss are comparable to those obtained with cross-entropy loss in all cases. We conduct all our experiments based on the code released in  \textcolor{blue}{\url{https://github.com/wyharveychen/CloserLookFewShot}} and \textcolor{blue}{\url{https://github.com/kjunelee/MetaOptNet}}.

\end{document}